\documentclass{article} 

\usepackage{microtype}
\usepackage{booktabs}
\usepackage{multirow}
\usepackage[]{algorithm2e}

\usepackage{hyperref}

\usepackage[preprint]{icml2020}

\icmltitlerunning{The Hidden Uncertainty in a Neural Network’s Activations}

\usepackage{times}

\usepackage{amsmath,amsfonts,bm}

\def\eqref#1{equation~\ref{#1}}

\def\1{\bm{1}}

\def\vx{{\bm{x}}}

\DeclareMathAlphabet{\mathsfit}{\encodingdefault}{\sfdefault}{m}{sl}
\SetMathAlphabet{\mathsfit}{bold}{\encodingdefault}{\sfdefault}{bx}{n}

\def\gL{{\mathcal{L}}}

\usepackage{hyperref}
\usepackage{url}

\usepackage{caption}
\usepackage{subcaption}
\usepackage{wrapfig}
\usepackage{tikz}
\usepackage{makecell}

\usepackage{blindtext}

\usepackage{graphicx}
\usepackage{acro}
\DeclareAcronym{mdl}{
  short = MDL ,
  long = minimum description length
}
\DeclareAcronym{ood}{
  short = OOD ,
  long = out-of-distribution
}
\DeclareAcronym{bdl}{
  short = BDL ,
  long = Bayesian Deep Learning
}
\DeclareAcronym{gmm}{
  short = GMM ,
  long = Gaussian mixture model
}
\DeclareAcronym{auroc}{
  short = AUROC ,
  long = area under the ROC curve
}
\DeclareAcronym{ml}{
  short = ML,
  long = machine learning
}
\DeclareAcronym{cnf}{
  short = CNF,
  long = conditional normalizing flow
}
\DeclareAcronym{pca}{
  short = PCA,
  long = principal component analysis
}
\DeclareAcronym{nfs}{
  short = NFs,
  long = Normalizing Flows
}
\DeclareAcronym{gp}{
  short = GP,
  long = Gaussian process
}

\begin{document}

\twocolumn[
\icmltitle{The Hidden Uncertainty in a Neural Network’s Activations}



\icmlsetsymbol{equal}{*}

\begin{icmlauthorlist}
\icmlauthor{Janis Postels}{equal,eth}
\icmlauthor{Hermann Blum}{equal,eth}
\icmlauthor{Yannick Strümpler}{eth}
\icmlauthor{Cesar Cadena}{eth}
\icmlauthor{Roland Siegwart}{eth}
\icmlauthor{Luc Van Gool}{eth}
\icmlauthor{Federico Tombari}{goo,tum}
\end{icmlauthorlist}

\icmlaffiliation{eth}{ETH Zurich}
\icmlaffiliation{goo}{Google}
\icmlaffiliation{tum}{Technical University Munich}

\icmlcorrespondingauthor{Janis Postels}{jpostels@ethz.ch}
\icmlcorrespondingauthor{Hermann Blum}{blumh@ethz.ch}

\icmlkeywords{Machine Learning, ICML}

\vskip 0.3in
]



\printAffiliationsAndNotice{\icmlEqualContribution} 


\begin{abstract}
The distribution of a neural network's latent representations has been successfully used to detect \ac{ood} data. This work investigates whether this distribution moreover correlates with a model's epistemic uncertainty, thus indicates its ability to generalise to novel inputs. We first empirically verify that epistemic uncertainty can be identified with the surprise, thus the negative log-likelihood, of observing a particular latent representation. Moreover, we demonstrate that the output-conditional distribution of hidden representations also allows quantifying aleatoric uncertainty via the entropy of the predictive distribution. We analyse epistemic and aleatoric uncertainty inferred from the representations of different layers and conclude that deeper layers lead to uncertainty with similar behaviour as established - but computationally more expensive - methods (e.g. deep ensembles). While our approach does not require modifying the training process, we follow prior work and experiment with an additional regularising loss that increases the information in the latent representations. We find that this leads to improved \ac{ood} detection of epistemic uncertainty at the cost of ambiguous calibration close to the data distribution. We verify our findings on both classification and regression models.
\end{abstract}

\section{Introduction}\label{chapter:introduction}

The recent success of deep neural networks in a variety of applications \cite{hinton2012speech,Redmon2016-qx} has been primarily driven by improvements in predictive performance, while progress on safety-related issues remained relatively slow \cite{amodei2016concrete,bozhinoski2019safety,janai2017computer}. Thereby, quantifying the uncertainty associated with the predictions of a neural network is a central challenge and a prerequisite for their broader deployment. One commonly distinguishes between two types of uncertainty \cite{Kiureghian2009-co}. Uncertainties arising from noise in the data regardless of the model are called \emph{aleatoric}. Uncertainties originating from the model choice and parameter fitting are referred to as \emph{epistemic}. This differentiation also has practical reasons, since epistemic uncertainty is reducible - e.g. by using more data -  while aleatoric uncertainty, as a property of the data, is not.

The predominant holistic framework for both aleatoric and epistemic uncertainty is \ac{bdl} \cite{mackay1992practical,hinton1993keeping}. However, scaling BDL to neural networks of practical size remains an open research question. Although recent research proposed various scalable approaches \cite{kingma2015variational,gal2016dropout,zhang2017noisy,postels2019sampling}, they still fall short of delivering a widely adopted practical solution. Moreover, the quality of the resulting posterior distribution, in particular for scalable approximations, is open to debate \cite{osband2016risk, wenzel2020good}. 

Prior work used the hidden representations for detecting \ac{ood} samples \cite{papernot2018deep, lee2018simple}. Moreover, density estimates of latent distributions have been implied to yield epistemic uncertainty when regularised appropriately \cite{alemi2018uncertainty, van2020simple}. The prospect that the distribution of latent representations holds information about the uncertainty associated with the predictions of a neural network promises efficient and scalable uncertainty estimates. However, prior work enforced additional constraints on the architecture, focused on classification and only evaluated on \ac{ood} detection. Nevertheless, epistemic uncertainty should not only identify far-away \ac{ood} samples, but also indicate how well a model generalises to its input, thus correlate with model performance in proximity of the training data distribution. Furthermore, hidden representations have, thus far, not been shown to also quantify aleatoric uncertainty. We demonstrate the latter by estimating the output-conditional density.

We show that the latent distribution in fact gives rise to a holistic framework for estimating both epistemic and aleatoric uncertainty for classification and regression - \textit{even without additional regularisation of the latent space} enabling widespread application. To this end, we examine uncertainty estimates extracted from different architectures and layers of various depth. Aleatoric uncertainty is assessed by its correlation with the prediction error on the training data distribution. Epistemic uncertainty is evaluated using \ac{ood} data - far away from and close to the training data distribution. We find that density estimates of shallow layers yield more conservative epistemic uncertainty, i.e. are more likely to label new data as \ac{ood}. The density of deeper layers behaves less conservatively and more similar to established - but computationally more expensive - epistemic uncertainty estimates (section \ref{experiments:image_classification}). Moreover, we discuss practical challenges that arise from uncertainty estimation based on hidden representations (section \ref{theory:pitfalls}). We further investigate how additional regularisation at training time in our framework impacts epistemic uncertainty (section \ref{experiments:reconstruction}).



\section{Related Work}\label{chapter:realted_work}

\subsection{Density Estimation for Out-of-Distribution Detection}\label{realted_work:ood_detection}

Density estimation is an established technique for OOD detection. While research also developed alternative methods \cite{pimentel2014review,chalapathy2019deep}, generative models are particularly appealing, since they denote a principled, unsupervised approach to the problem by learning the data distribution \cite{schlegl2017unsupervised,li2018anomaly}. However, concerns have been voiced regarding their effectiveness \cite{vskvara2018generative} and, in particular for explicit generative models, vulnerability to OOD samples \cite{Nalisnick2018-gd}. Ensembles of generative models \cite{Choi2018-lm} have been proposed to solve the latter, as well as considering the distribution of log-likelihoods \cite{nalisnick2019detecting, morningstar2020density}.

Since learning high-dimensional densities is challenging, another line of research focuses on applying machine learning approaches to the latent space of neural networks \cite{ruff2018deep,papernot2018deep} or directly estimating its density, which is often simpler than for the data itself. \cite{lee2018simple} fits a~\ac{gmm} to the representations of a neural network. \cite{Blum2019-eh} trains a normalizing flow on the latent space. Moreover, \cite{alemi2018uncertainty} utilizes the rate term in the Information Bottleneck (VIB) \cite{alemi2016deep} for OOD detection.

\vspace{-2mm}

\subsection{Uncertainty Estimation}\label{realted_work:uncertainty_estimation}

 The idea that uncertainty should increase outside of the training data distribution relates uncertainty estimation to \ac{ood} detection. Thus, uncertainty estimation has been applied to the latter \cite{hendrycks2016baseline, li2017dropout, snoek2019can}. We discuss this relation in section
~\ref{theory:uncertainties_in_ml}.
 
\ac{bdl} \cite{neal2012bayesian,neal2011mcmc,hinton1993keeping} constitutes the predominant way to quantify uncertainty, since it handles both aleatoric and epistemic uncertainty in a principled manner. Due to well known difficulties of applying \ac{bdl} to large neural networks, numerous attempts have been proposed \cite{kingma2015variational,gal2016dropout,zhang2017noisy, teye2018bayesian,postels2019sampling}. However, current approaches still lack scalability or quality of the predicted uncertainty.
 
 A parallel line of work uses the predictive conditional distribution of neural networks to quantify uncertainty \cite{bishop1994mixture,hendrycks2016baseline,lee2017training}. However, these approaches meet fundamental difficulties outside of the training data distribution. Furthermore, neural networks have been found to be poorly calibrated \cite{guo2017calibration}, which poses a challenge to all uncertainty estimates.
 
 Recently, it has been suggested that the distribution of a neural network's latent representations quantifies epistemic uncertainty. \cite{Mandelbaum2017-ti} fits it with a kernel density. They outperform the softmax entropy and Monte-Carlo (MC) dropout \cite{gal2016dropout} when imposing an additional regularising loss on the latent space. \cite{van2020simple} replaces the softmax activation with a target-conditional Gaussian distribution with fixed variance for classification. Their approach paired with a gradient penalty on the input, boosts \ac{ood} performance by enforcing distance awareness of the representations. They refer to the distance to the closest target-conditional mean as epistemic uncertainty and evaluate it on OOD detection. Similarly, \cite{liu2020simple} enforces the representations of a neural network to be distance-aware by constraining its Lipschitz constant. In \cite{alemi2018uncertainty} the authors consider the per-sample rate loss in the VIB objective \cite{alemi2016deep} and the softmax entropy as proxies for uncertainty. They show that the softmax entropy excels at quantifying aleatoric uncertainty, and the rate-loss at OOD detection. Above works are united by the fact that they introduce additional regularisation which forces the hidden representations to store more information about the input. Additional hyperparameters are tuned to maximise \ac{ood} performance. We discuss this further in section \ref{theory:pitfalls}. On the contrary, our work reveals that similar conclusions can be drawn for deterministic neural networks without altering the training procedure while additionally incorporating aleatoric uncertainty.
\vspace{-2mm}
\section{Uncertainty from Latent Representations}\label{chapter:theory}%

\subsection{Types of Uncertainty}\label{theory:uncertainties_in_ml}

Uncertainty estimation in \ac{ml} tries to assign a level of confidence to a model's output. Thus, the true uncertainty correlates with a model's performance. While a correct prediction can have high uncertainty, average model performance degrades when uncertainty increases.

A model's performance fluctuates within the training data distribution. One reason for this is inherent noise in the data, which causes irreducible \emph{aleatoric} uncertainty~\cite{Kiureghian2009-co}. Additionally, the performance depends on the choice of training data, model, and parameters. Uncertainties and performance drops resulting from these choices are reducible and should be distinguished from their irreducible counterparts. \cite{Kiureghian2009-co} call these uncertainties `\emph{epistemic}'. The same terminology was introduced to the \ac{ml} community~\cite{senge2014reliable, gal2016uncertainty}. It is worth noting that some works aim at separating \emph{distributional} from \emph{epistemic} uncertainty \cite{malinin2018predictive,harang2018towards}. We follow the definitions from~\cite{Kiureghian2009-co}, where \emph{epistemic} uncertainty encompasses all reducible uncertainty.

\vspace{-4mm}

\paragraph{The Two Faces of Epistemic Uncertainty.}\label{theory:two_faces_of_epistemic_uncertainty.} Many related works solely evaluate epistemic uncertainty using \ac{ood} detection \cite{Mandelbaum2017-ti,alemi2018uncertainty,van2020simple,liu2020simple}. This wrongly equates epistemic uncertainty estimation with \ac{ood} detection. While \ac{ood} detection, as an important failure source, is related to epistemic uncertainty estimation and, thus, is a necessary evaluation method, we stress that \ac{ood} detection performance alone is not sufficient. This becomes apparent when considering that \ac{ood} detection represents a model agnostic task and epistemic uncertainty should be aligned with a model's generalisation. Consequently, high epistemic uncertainty is only expected for those \ac{ood} samples that lead to degradation of model performance. In this work we therefore evaluate epistemic uncertainty along two dimensions - its behaviour close to and far away from the training distribution. We expect that a good epistemic uncertainty estimate successfully detects samples that reside far away from the data distribution as well as indicates how well the model generalises to data close to it. We particularly investigate the latter in section \ref{experiments:image_classification} by applying perturbations of increasing magnitude to the input while tracking estimated uncertainty and model performance.


\subsection{Uncertainty of Deterministic Neural Networks}\label{theory:uncertainties_in_deterministic_nn}

We now establish a theoretical connection between the density of hidden representations and the information-theoretic surprise of observing a particular sample or prediction under the assumption of a deterministic neural network. While we require the model to be deterministic here, this only effects the inference stage of a model. Thus, one is free to use any stochastic regularisation methods at training time. Moreover, we relate this surprise to aleatoric and epistemic uncertainty estimates used by prior work.

Given a dataset $\mathcal{D}=\{ \textbf{X}, \textbf{Y} \}$, the goal of a discriminative task is to model the conditional distribution $p(\textbf{y}|\textbf{x})$, where $\textbf{x} \in \textbf{X}$ and $\textbf{y} \in \textbf{Y}$. Let $\hat{\textbf{y}}$ correspond to the output of a neural network trained to approximate $p(\textbf{y}|\textbf{x})$, which induces the  conditional distribution $p_{\theta}(\hat{\textbf{y}}|\textbf{x})$. The conditional entropy 
\begin{align}\label{theory:conditional_entropy}
H(\hat{\textbf{y}}|\textbf{x}) &= E_{p_{\theta}(\hat{\textbf{y}},\textbf{x})} \left[ -\log \left( p_{\theta}(\hat{\textbf{y}}|\textbf{x}) \right)  \right] \notag \\
&= E_{p(\textbf{x})} \left[ -\int \mathrm{d}\hat{\textbf{y}}  p_{\theta}(\hat{\textbf{y}}|\textbf{x}) \log \left( p_{\theta}(\hat{\textbf{y}}|\textbf{x}) \right)  \right]
\end{align}
corresponds to the total uncertainty about $\hat{\textbf{y}}$ given knowledge of $\textbf{x}$ across the dataset $\mathcal{D}$. Both quantities under the expectation have been successfully used to quantify aleatoric uncertainty in deterministic neural networks for classification \cite{hendrycks2016baseline,alemi2018uncertainty}\footnote{In fact, \cite{hendrycks2016baseline} uses the maximum softmax probability. This does not change the ordering of the uncertainty values, since the logarithm is a strictly monotonic function.
} and regression \cite{kendall2017uncertainties}\footnote{The authors parameterize the output with a unimodal Gaussian and use its variance as aleatoric uncertainty. In this case the entropy is proportional to the logarithm of the variance.}. The negative log-likelihood $-\log(p_{\theta}(\hat{\textbf{y}}|\textbf{x}))$ is called surprisal of a particular value $\hat{\textbf{y}}$ given an input $\textbf{x}$. Further, $-\int \mathrm{d}\hat{\textbf{y}}  p_{\theta}(\hat{\textbf{y}}|\textbf{x}) \log \left( p_{\theta}(\hat{\textbf{y}}|\textbf{x})  \right)$ is the expected surprisal for a given input $\textbf{x}$.

However, eq. \ref{theory:conditional_entropy} only quantifies the uncertainty/surprise about predictions given a specific input $\textbf{x}$ and not about observing $\textbf{x}$ in the first place. To account for this, consider the entropy of the joint probability distribution 
\begin{align}\label{theory:joint_entropy}
&H(\hat{\textbf{y}}, \textbf{x}) = H(\textbf{x}) + H(\hat{\textbf{y}}|\textbf{x}) \\
&= E_{p(\textbf{x})} \left[ -\log(p(\textbf{x})) - \int \mathrm{d}\hat{\textbf{y}}  p_{\theta}(\hat{\textbf{y}}|\textbf{x}) \log \left( p_{\theta}(\hat{\textbf{y}}|\textbf{x}) \right)  \right]. \notag
\end{align}
In eq. \ref{theory:joint_entropy} the negative log-likelihood $-\log(p(\textbf{x}))$ quantifies the surprise about observing $\textbf{x}$ and requires learning the density $p(\textbf{x})$. Estimating $-\log(p(\textbf{x}))$ directly would provide an uncertainty estimate independent of the neural network used to parameterize $p_{\theta}(\hat{\textbf{y}}|\textbf{x})$ and its generalisation. It can be considered as a model-agnostic distributional uncertainty. Moreover, this distributional uncertainty incorporates the entire information in the distribution of $\textbf{X}$ - also information irrelevant for the discriminative task. Additionally, a high-dimensional $\textbf{X}$ induces further complications, since learning such densities is an active field of research. 

Consider a neural network comprised of $L$ layers with $L-1$ latent representations $(\textbf{z}_0, .., \textbf{z}_{L-2})$ with a joint distribution factorizing according to $p_{\theta}(\textbf{x}, \hat{\textbf{y}}, \textbf{z}_0, .., \textbf{z}_{L-2}) = p_{\theta}(\hat{\textbf{y}}|\textbf{z}_{L-2}) p(\textbf{z}_{L-2}|\textbf{z}_{L-3})...p_{\theta}(\textbf{z}_0|\textbf{x})p(\textbf{x})$. Exploiting its deterministic nature and the data processing inequality allows us to make the following statements:
\begin{gather}\label{theory:entropy_inequalities}
    H(\hat{\textbf{y}}|\textbf{z}_i) = H(\hat{\textbf{y}}|\textbf{x}) \quad \forall i \in [0, ..., L-2]\\
    H(\textbf{x}) \geq H(\textbf{z}_{0}) \geq ... \geq H(\textbf{z}_{L-2}) \label{eq:hz}
\end{gather}%
A detailed derivation is in the supplement. These inequalities relate the distribution of the latent representations to the uncertainties in eq.~\ref{theory:joint_entropy}. Inequality \ref{eq:hz} implies that uncertainty estimates based on deeper layers are less conservative. On the other side, since the distribution of the latent space at a particular layer is a functional of all the previous layers, we expect these uncertainty estimates to incorporate some degree of model dependence. We verify this intuition empirically in section \ref{experiments:image_classification} by detecting \ac{ood} samples using density estimate of different layers.

We also investigate whether the latent distribution contains information about aleatoric uncertainty. Therefore, we learn the joint density $p(\textbf{z}_i, \hat{\textbf{y}})$ at layer $i$. This can be achieved by separately estimating the output-conditional distribution $p(\textbf{z}_i|\hat{\textbf{y}})$ and $p(\hat{\textbf{y}})$. Using Bayes' formula and marginalization over $\hat{\textbf{y}}$, we can compute both values under the expectations in eq.~\ref{theory:conditional_entropy}. In section \ref{experiments:image_classification} we find empirically that deeper layers enable better aleatoric uncertainty estimates. We hypothesize that this results from the neural network filtering out task-irrelevant information, which makes it easier to estimate the output-conditional density. 

To summarise, given a novel input $\textbf{x}^\star$ with corresponding latent vector $\textbf{z}_i^\star$, we identify the epistemic uncertainty with the surprise of observing that latent representation $-\log p(\textbf{z}_i^\star)$ and the aleatoric uncertainty with the expected surprisal of $\hat{\textbf{y}} \sim p(\hat{\textbf{y}}|\textbf{z}_i^\star)$ which we subsequently denote with $h(\hat{\textbf{y}}|\textbf{z}_i^\star)$:
\begin{align}
    -\log p(\textbf{z}^\star) &= -\log \left( \int \mathrm{d}\hat{\textbf{y}} p(\textbf{z}^\star|\hat{\textbf{y}}) p(\hat{\textbf{y}}) \right)\label{eq:epistemic} \\
    h(\hat{\textbf{y}}|\textbf{z}_i^\star) &=  -\int \mathrm{d}\hat{\textbf{y}}  p_{\theta}(\hat{\textbf{y}}|\textbf{z}_i^\star) \log \left( p_{\theta}(\hat{\textbf{y}}|\textbf{z}_i^\star) \right) \label{eq:aleatoric}
\end{align}%

\subsection{Pitfalls}\label{theory:pitfalls}

\textbf{Feature Collapse.} One of the strengths of discriminative neural networks is to ignore task-irrelevant features of the data. However, this can lead to activations of \ac{ood} data becoming indistinguishable from activations of the training data. This phenomenon is also called feature collapse~\cite{van2020simple} and poses a practical challenge for uncertainty quantification based on hidden activations. Different prior works have addressed this problem by tuning additional hyperparameters enforcing adequately structured latent spaces. Hereby, the network is made more sensitive to changes in the input by constraining its Lipschitz constant. This is achieved by using a distance-based loss on the hidden representations \cite{Mandelbaum2017-ti}, a gradient penalty \cite{van2020simple} or spectral normalization \cite{liu2020simple}. Further, although \cite{alemi2018uncertainty} does not mention it explicitly, training with the VIB framework regularises the distribution of hidden representations according to a prior distribution. 

In our main contribution (section \ref{experiments:image_classification} \& \ref{experiments:regression}) we do not explicitly address feature collapse, since it requires altering the training procedure, adds architectural constraints and, most importantly, risks overfitting epistemic uncertainty to the task of \ac{ood} detection. All mentioned prior works focus on \ac{ood} detection and it remains unclear whether such approaches translate into good epistemic uncertainty, especially close to the training data distribution. We demonstrate in section \ref{experiments:image_classification} that, \textit{without additional regularisation}, the density of the hidden representations of deeper layers as a proxy for epistemic uncertainty is similarly calibrated as other well established - but computationally more expensive - methods, e.g. deep ensembles \cite{lakshminarayanan2017simple}, while maintaining good \ac{ood} detection performance.

Furthermore, in order to link these results to prior works above and illustrate the risk of overfitting epistemic uncertainty to \ac{ood} detection, we explore within the framework of eq.~\ref{eq:hz} how regularisation techniques that yield $\textbf{z'}_l$ such that $H(\textbf{x}) \geq H(\textbf{z'}_l) \geq H(\textbf{z}_l)$ influence the estimated epistemic uncertainty. 
Therefore, we augment the original loss by a weighted reconstruction objective to achieve more informative latent representations $\textbf{z'}_l$:
\begin{equation}
 \Tilde{\gL} = \gL_{orig} + \lambda \, \text{MSE}(\vx, \hat{\vx})
\end{equation}
where $\vx$ is the input and $\hat{\vx} = D(\textbf{z'}_l)$ is the reconstruction from an additional decoder network $D$, which is a symmetric version of the encoding network (see supplement for details). We evaluate the impact of this regularisation on epistemic uncertainty in section \ref{experiments:reconstruction}.

\textbf{High-Dimensional Densities.} The hidden representations are high-dimensional for large neural networks or shallow layers. This renders estimating their density impractical. Prior work has also faced the curse of dimensionality. For example, \cite{lee2018simple} reduced the dimensionality of hidden representations by performing \ac{pca} on the training data and, subsequently, reusing the transformation at inference time. Moreover, \cite{papernot2018deep} uses locality-sensitive hashing in combination with random projections for nearest neighbour search. Here, we follow \cite{lee2018simple} applying \ac{pca} and find that it yields satisfying results in practice. However, we note that \ac{pca} as well as random projections are not optimal solutions in the case of convolutional neural networks whose hidden representations entail translational invariances that get lost by applying dense projection matrices.

\subsection{Implementation}\label{theory:implementation}

In order to then estimate epistemic (eq. \ref{eq:epistemic}) and aleatoric uncertainty (eq. \ref{eq:aleatoric}), we extract the latent representations of a particular layer $i$ and learn the output-conditional density $p(\textbf{z}_i|\hat{\textbf{y}})$ and the marginal distribution of the neural network's outputs $p(\hat{\textbf{y}})$ on the training data.

We learn $p(\textbf{z}_i|\hat{\textbf{y}})$ using an explicit generative model. For image classification (section \ref{experiments:image_classification}), we use one \ac{gmm} with $k \in \mathbb{N}_{\setminus \{0\}}$ components for each class. For regression (section \ref{experiments:regression}) which entails continuous predictions, we found that \acp{cnf} and in particular the conditioning scheme used in \cite{Ardizzone2019-bv} yielded good results. We refer the reader to the supplementary material for details.

Estimating $p(\hat{\textbf{y}})$ for a classification task simply involves counting the predicted labels on the training data. For a regression task, however, it is necessary to estimate the density of the predictions in the output space. This is relatively simple, since the output space is low dimensional for most common discriminative tasks. We found it sufficient to approximate it with common parametric distributions - a uniform distribution in section~\ref{experiments:regression} and a betaprime distribution for the predicted depth in the supplement.

We then obtain the epistemic uncertainty in eq.~\ref{eq:epistemic} by marginalizing $\hat{\textbf{y}}$. To solve the integral, we use summation (classification) or numerical integration (regression). We found the latter to be sufficient for the one dimensional output spaces in our regression experiments (see section~\ref{experiments:regression} and the supplement). Then, we calculate the aleatoric uncertainty in eq. \ref{eq:aleatoric} by applying Bayes' formula. 
\vspace{-2mm}%
\section{Experiments}\label{chapter:experiments}%
\begin{figure*}
\centering

\begin{subfigure}[b]{\linewidth}
\begin{minipage}{0.1\linewidth}
    \caption{}
\end{minipage}%
\begin{minipage}{0.3\linewidth}
    \includegraphics[width=\linewidth]{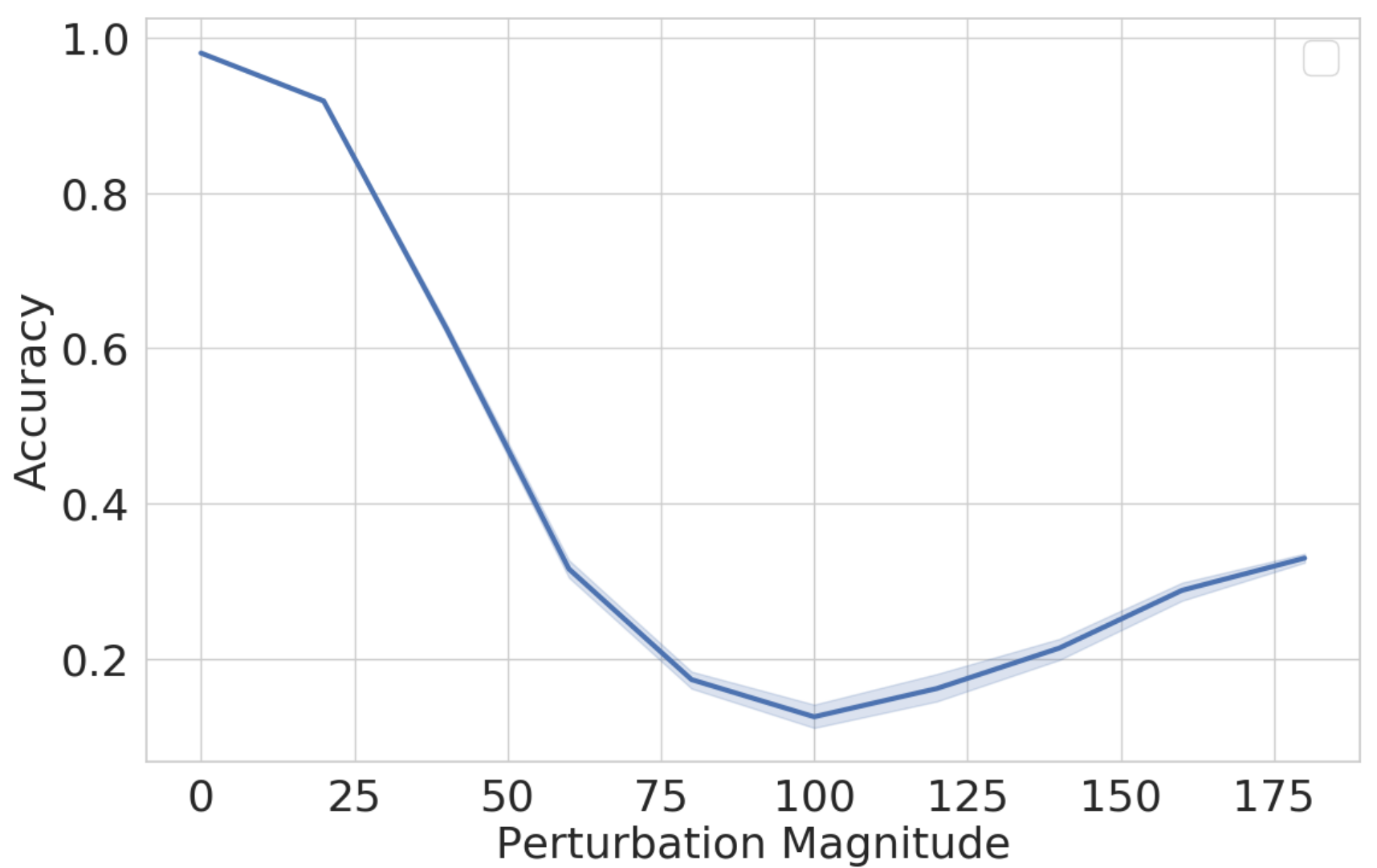}
\end{minipage}
\begin{minipage}{0.3\linewidth}
    \begin{subfigure}[b]{\linewidth}
        \includegraphics[width=\linewidth]{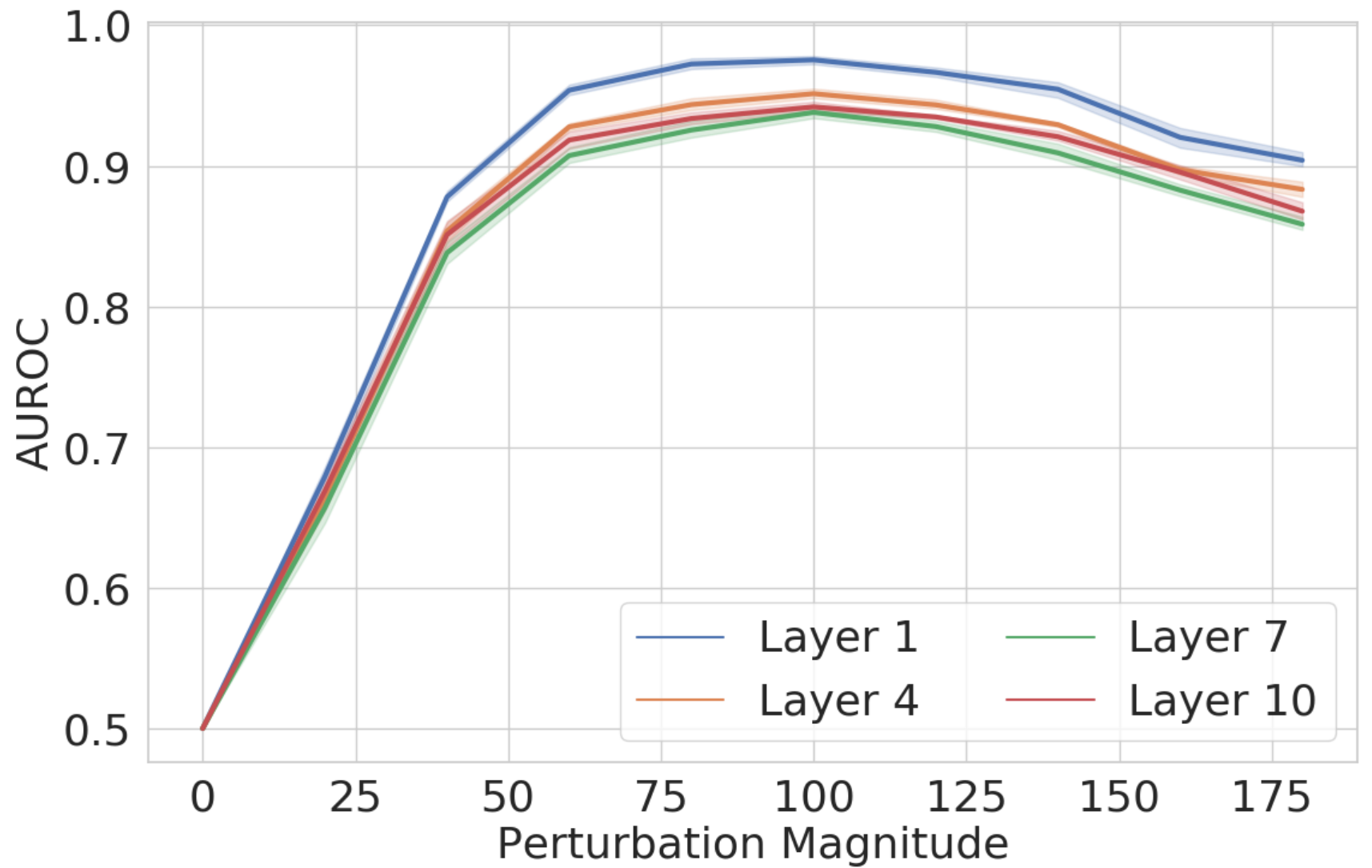}
    \end{subfigure}
\end{minipage}
\begin{minipage}{0.3\linewidth}
    \begin{subfigure}[b]{\linewidth}
        \includegraphics[width=\linewidth]{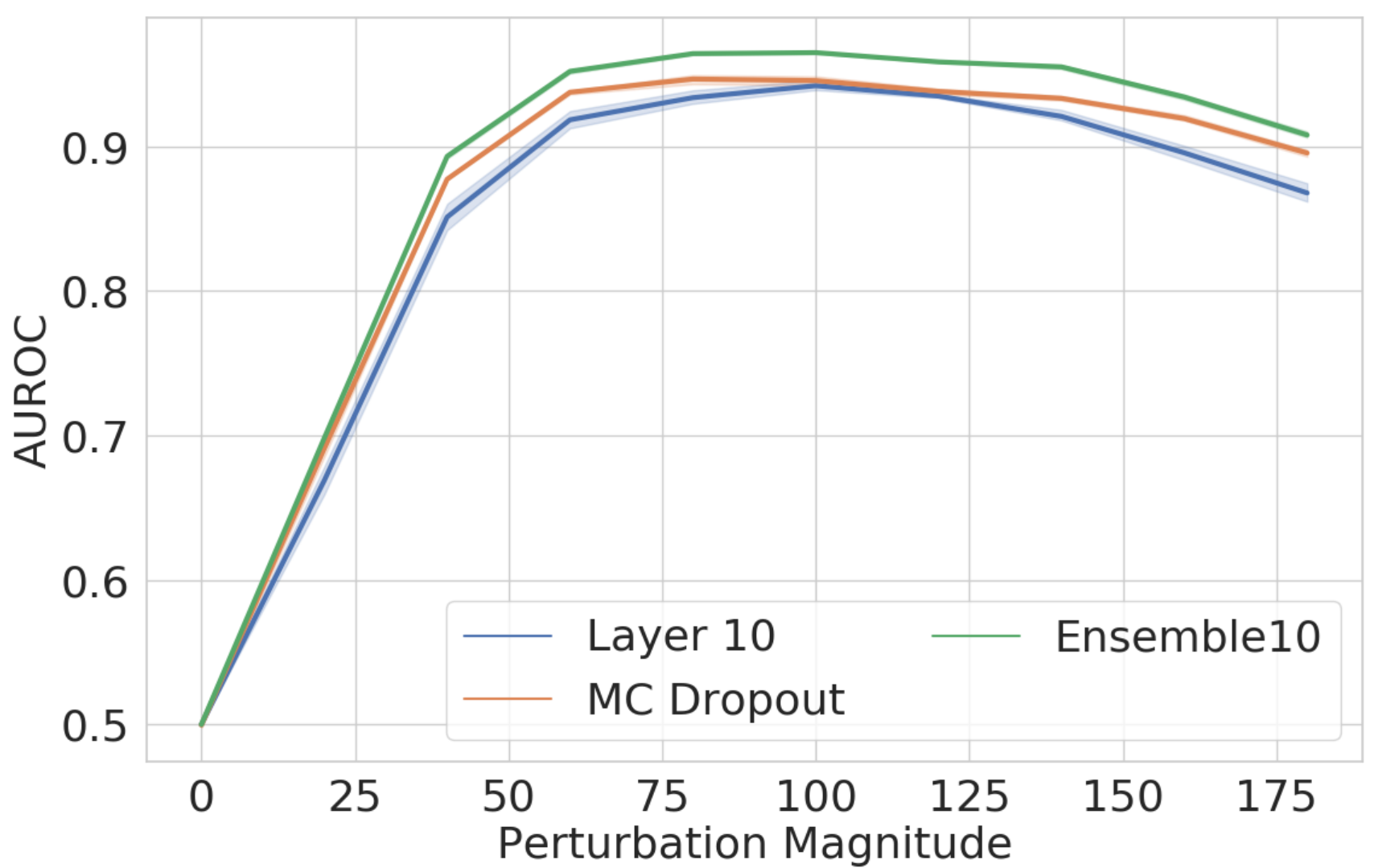}
    \end{subfigure}
\end{minipage}\hfill
\end{subfigure}

\begin{subfigure}[b]{\linewidth}
\begin{minipage}{0.1\linewidth}
    \caption{}
\end{minipage}%
\begin{minipage}{0.3\linewidth}
    \begin{subfigure}[b]{\linewidth}
        \includegraphics[width=\linewidth]{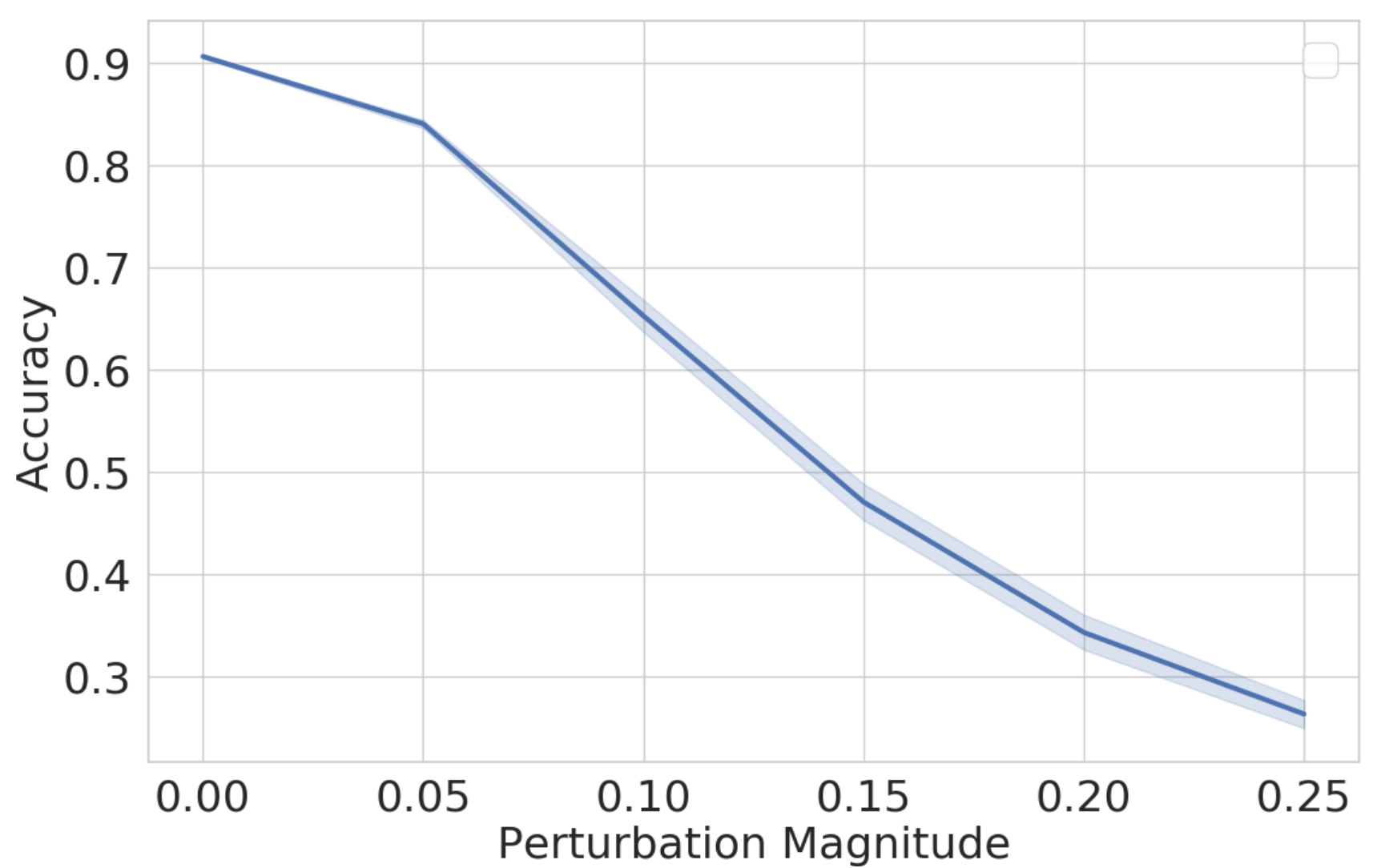}
    \end{subfigure}
    \caption*{Classification accuracy.}
\end{minipage}
\begin{minipage}{0.3\linewidth}
    \begin{subfigure}[b]{\linewidth}
        \includegraphics[width=\linewidth]{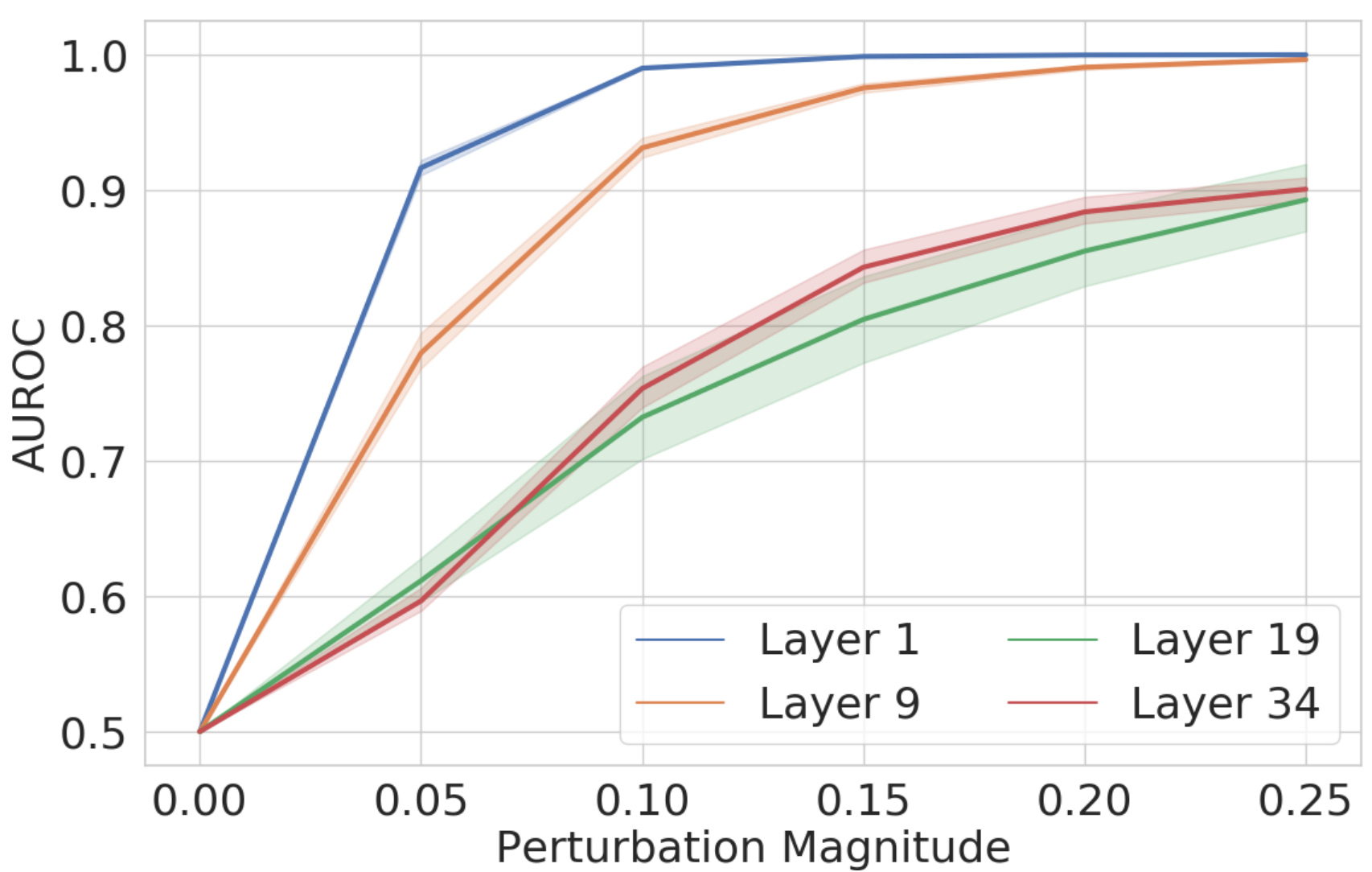}
    \end{subfigure}
    \caption*{Comparison of different layers.}
\end{minipage}
\begin{minipage}{0.3\linewidth}
    \begin{subfigure}[b]{\linewidth}
        \includegraphics[width=\linewidth]{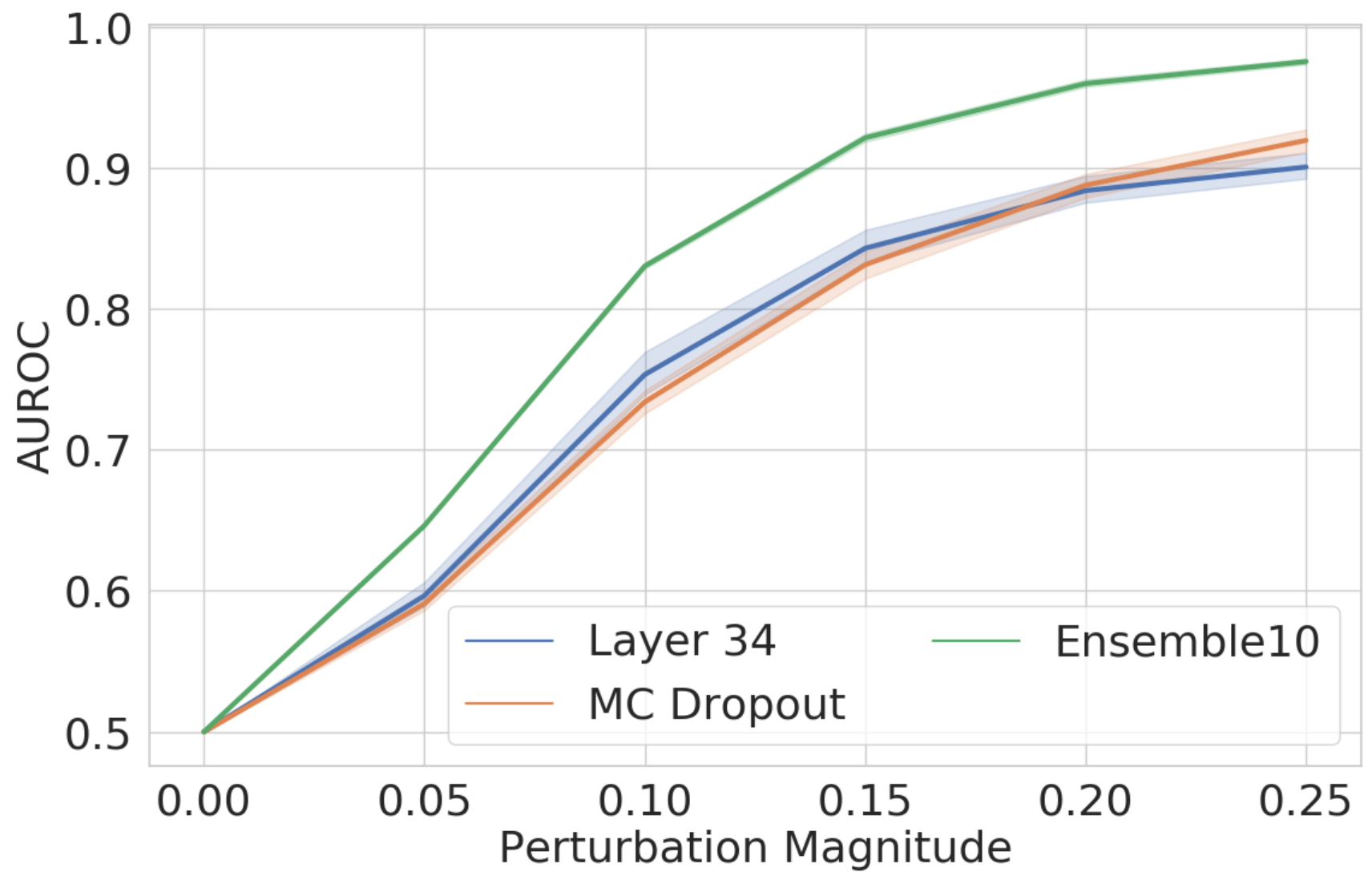}
    \end{subfigure}
    \caption*{Comparison with other methods.}
\end{minipage}\hfill
\end{subfigure}
\vspace{-2mm}
\caption{Analysis of the quality of estimated epistemic uncertainty over different layers for FCNet on MNIST (\textbf{a}), and ResNet18 on SVHN (\textbf{b}). We plot the accuracy of the classification network (\textbf{left}) and the \ac{auroc} (obtained using epistemic uncertainty as threshold) against the perturbation magnitude on the test data (\textbf{center} and \textbf{right}). We compare epistemic uncertainty obtained from hidden representations of varying depth (\textbf{center}) and other methods (\textbf{right}). We observe that (i) density estimates based on deeper layers behave similar as other established - but computationally more expensive - methods (i.e. MC Dropout and Deep Ensembles) and (ii) density estimates based on shallow layers yield more conservative estimates.}
\label{fig:qualitative_uncertainty}
\vspace{-4mm}
\end{figure*}

While there exists no ground truth for uncertainty, we follow the approach of other works~\cite{gal2016dropout,lakshminarayanan2017simple,malinin2018predictive} and examine epistemic uncertainty over input transformations and distributional shifts and aleatoric uncertainty via the correlation with the network's predictive performance on in-distribution data. We first investigate the behaviour of uncertainties extracted from the output-conditional density over varying depth of the representations on image classification. We then validate this approach for the task of regression on a toy dataset and finally investigate whether additional training losses can improve the extracted uncertainty. We refer to the supplementary material for additional experiments with semantic segmentation and depth regression, as well as all experimental details and parameters.

\subsection{Image Classification}\label{experiments:image_classification}

We train a multilayer perceptron (FCNet) consisting of fully-connected layers and ReLU activations on MNIST \cite{lecun1998gradient} and FashionMNIST \cite{xiao2017fashion} and ResNet18 \cite{he2016resnet} on CIFAR10 \cite{krizhevsky2009learning} and SVHN \cite{netzer2011reading}.

\begin{figure*}[t]
\centering
\begin{minipage}{0.245\linewidth}
    \begin{subfigure}[b]{\linewidth}
        \includegraphics[width=\linewidth]{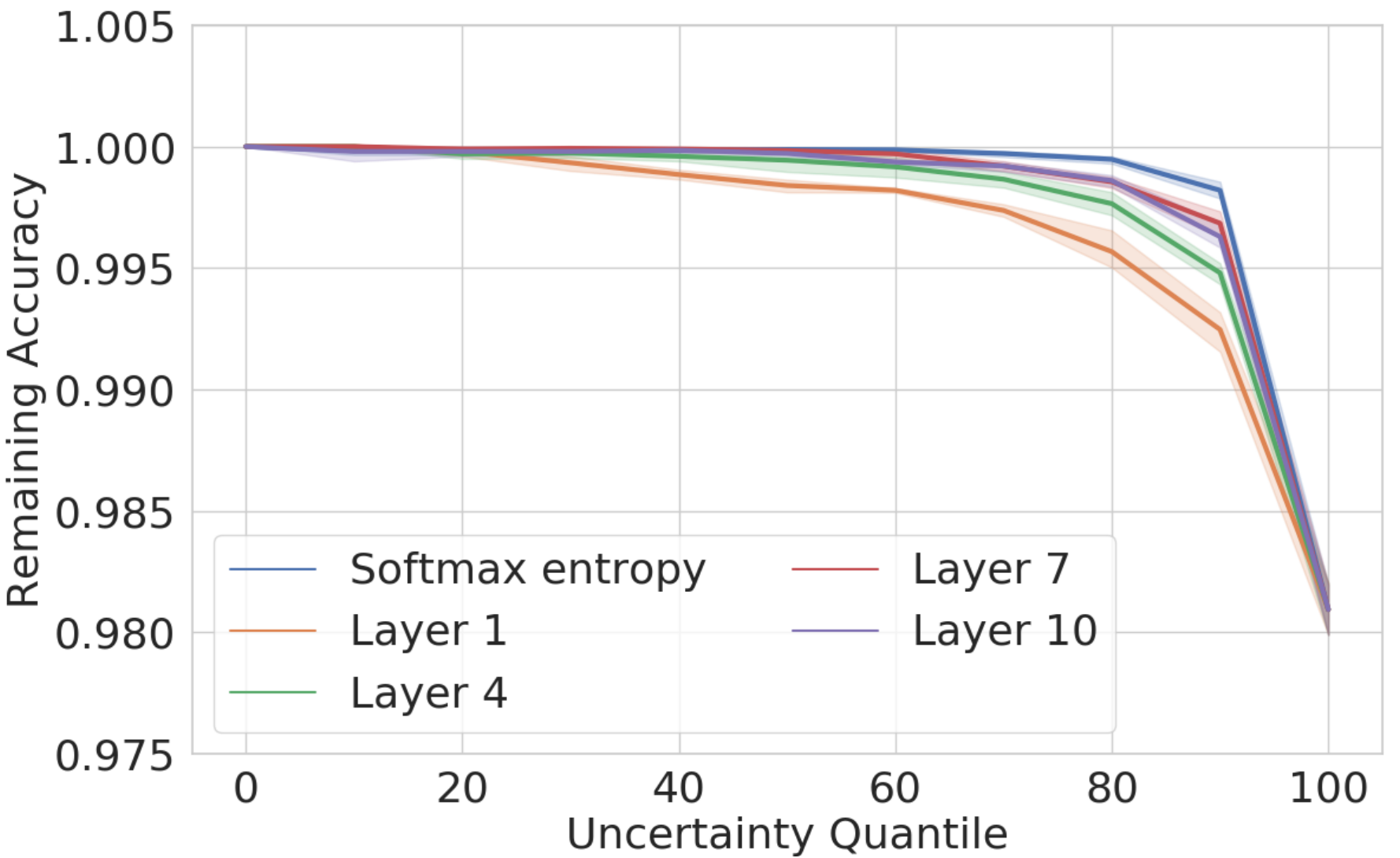}
    \end{subfigure}
\end{minipage}
\begin{minipage}{0.245\linewidth}
    \begin{subfigure}[b]{\linewidth}
        \includegraphics[width=\linewidth]{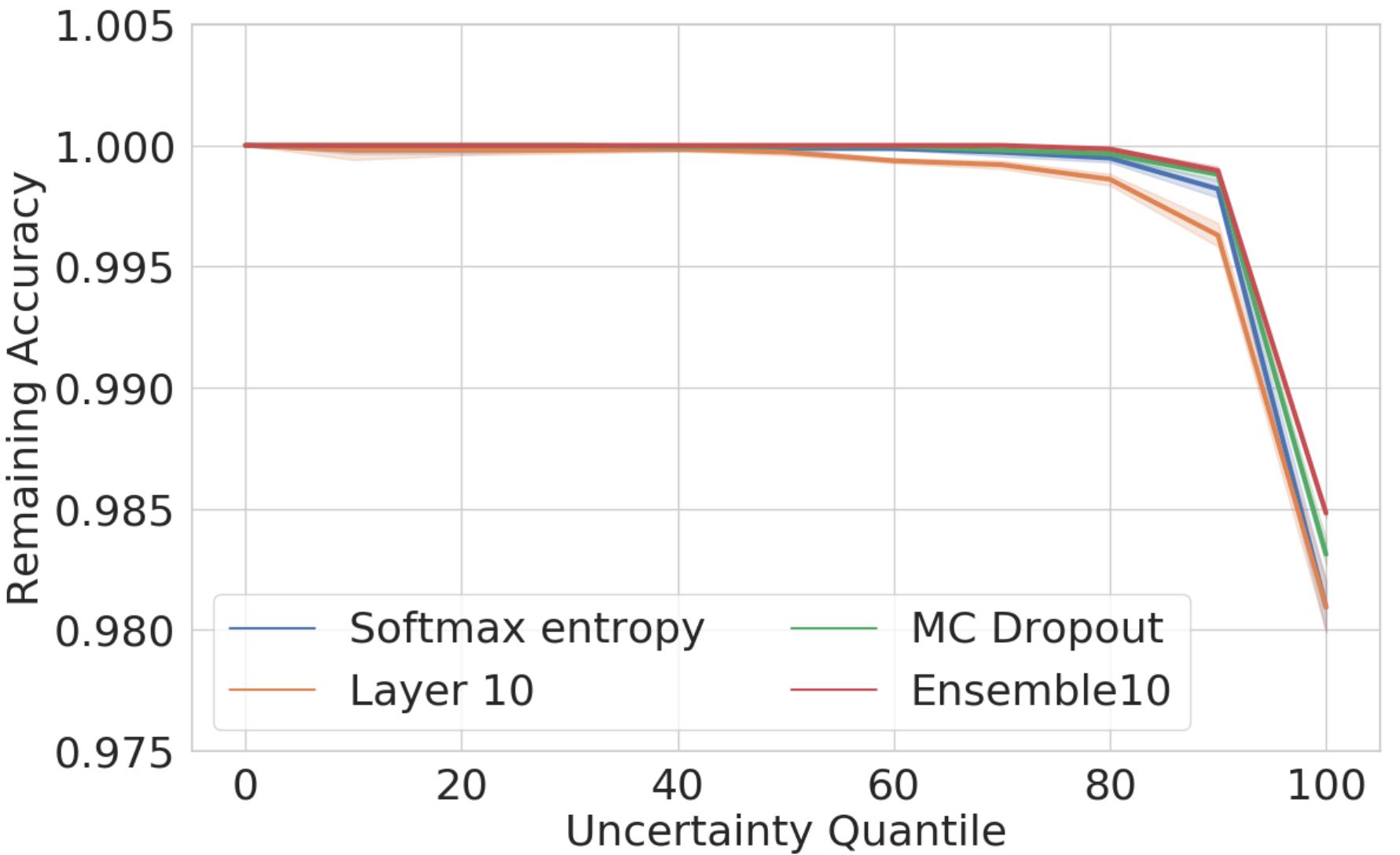}
    \end{subfigure}
\end{minipage}
\begin{minipage}{0.245\linewidth}
    \begin{subfigure}[b]{\linewidth}
        \includegraphics[width=\linewidth]{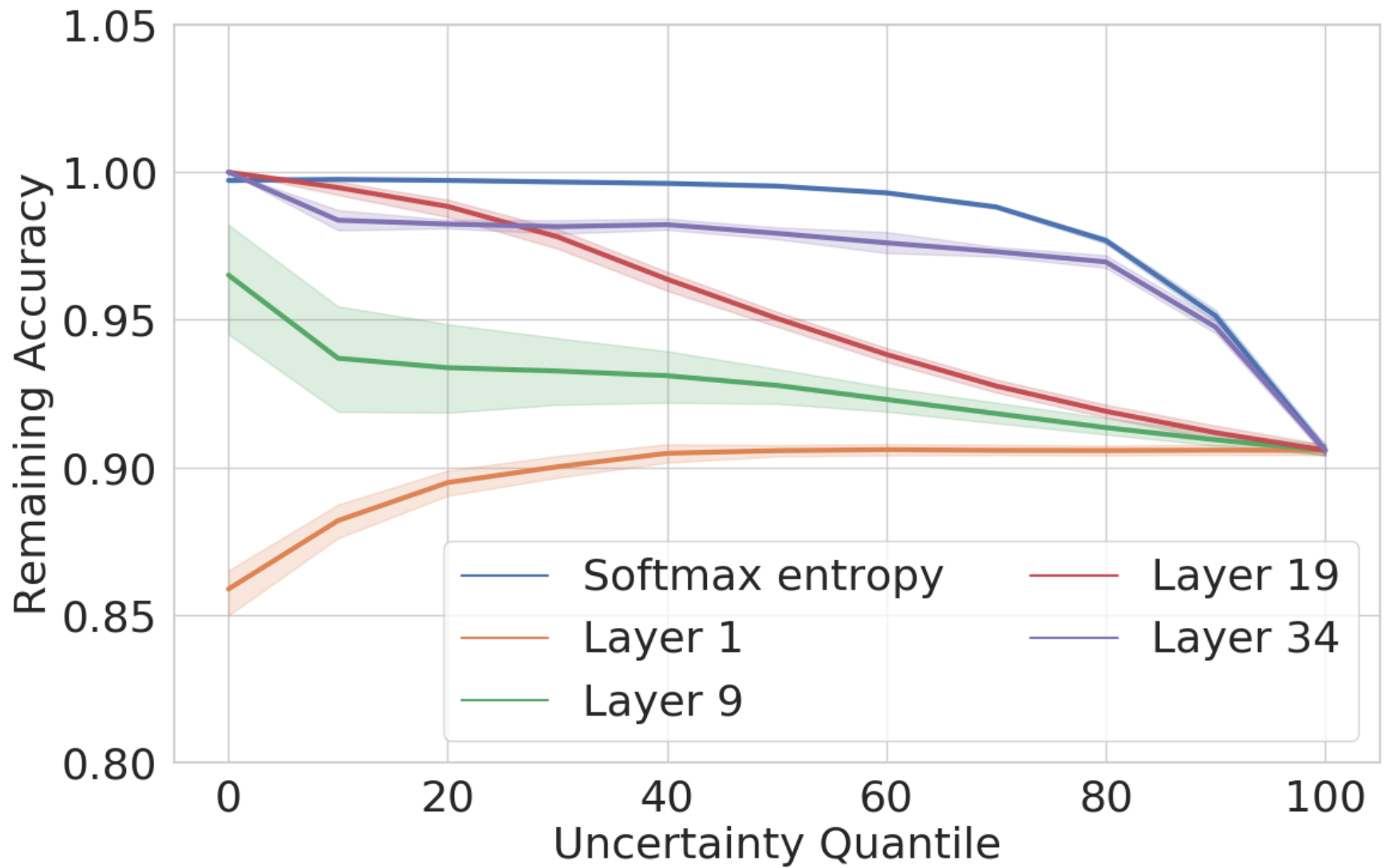}
    \end{subfigure}
\end{minipage}
\begin{minipage}{0.245\linewidth}
    \begin{subfigure}[b]{\linewidth}
        \includegraphics[width=\linewidth]{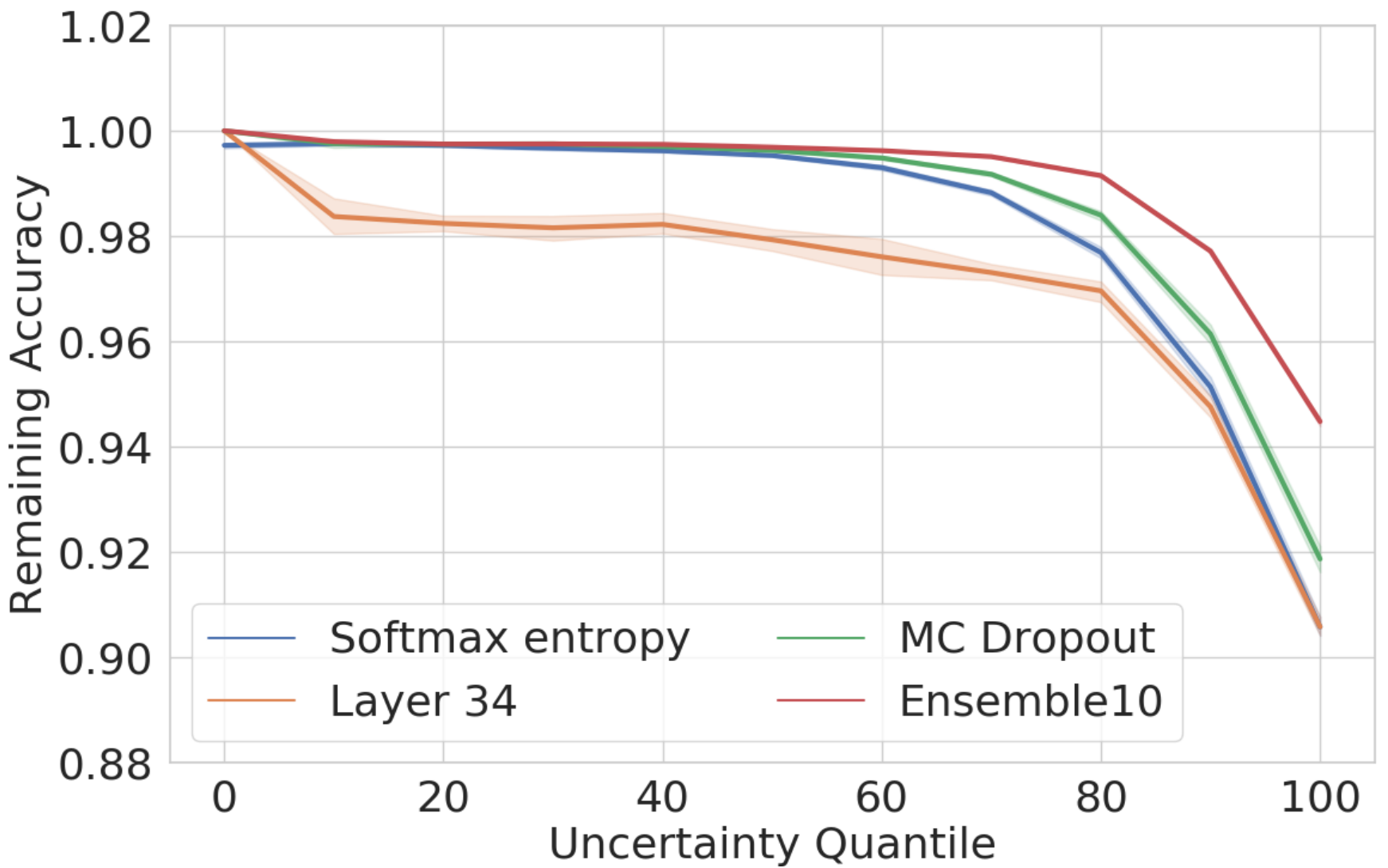}
    \end{subfigure}
\end{minipage}\hfill
\begin{minipage}{0.49\linewidth}
    \caption*{Aleatoric uncertainty on MNIST}
\end{minipage}
\begin{minipage}{0.49\linewidth}
    \caption*{Aleatoric uncertainty on SVHN}
\end{minipage}\hfill
\vspace{-5mm}
\caption{Analysis of the calibration of aleatoric uncertainty on MNIST and SVHN. We plot the remaining accuracy when selecting test samples using incremetally increased aleatoric uncertainty threshold. We observe that aleatoric uncertainty based on deeper layers tends to behave similar as other approaches (softmax entropy, deep ensembles, MC dropout).}
\label{fig:qualitative_uncertainty_calibration}
\vspace{-4mm}
\end{figure*}

\subsubsection{Depth Dependant Uncertainty Quality}\label{experiments:quality_of_uncertainty_depending_on_depth}

We estimate the output-conditional density of the hidden representations on the training data of FCNet trained on MNIST and of ResNet18 trained on SVHN using one \ac{gmm} with five components per class. Each training and subsequent density estimation is conducted five times independently yielding mean and standard deviation. We quantify epistemic and aleatoric uncertainty according to eq.~\ref{eq:epistemic} and eq.~\ref{eq:aleatoric}. We compare our results to MC dropout \cite{gal2016dropout} and deep ensembles \cite{lakshminarayanan2017simple} for which we compute aleatoric and epistemic uncertainty according to \cite{gal2017deep} (see supplement).

First, we evaluate the behaviour of the epistemic uncertainty estimate when the input is shifted away from the training data distribution in small discrete steps. We expect epistemic uncertainty to correlate with model generalisation, which is the observed behaviour of established methods such as MC dropout and deep ensembles. We chose to compare with these methods due to their popularity. We compute the \ac{auroc} between the unperturbed testset and each perturbation magnitude. The \ac{auroc} measures how well a binary classifier identifies the perturbed test samples based on the estimated epistemic uncertainty. On MNIST we rotate test samples from $0^{\circ}$ to $180^{\circ}$ in $20^{\circ}$ steps and on SVHN we add independent Gaussian noise to the pixel values of input images, while increasing the standard deviation from 0 to 80 with step size 10. Fig. \ref{fig:qualitative_uncertainty} shows the results.

We observe that the epistemic uncertainty obtained from density estimates from shallow layers behaves more conservatively in the sense that they are quicker to label perturbed data as \ac{ood}. On the other side, epistemic uncertainty estimates from deeper layers behave less conservatively and similar to MC Dropout and Deep Ensembles, while only requiring training one model and a single forward pass.

Fig. \ref{fig:qualitative_uncertainty_calibration} analyzes the quality of the estimated aleatoric uncertainty using calibration curves. These show the accuracy of the neural network depending on the uncertainty. We use a threshold sliding through increasing percentiles of the aleatoric uncertainty and plot the remaining accuracy. Ideally this yields a strictly monotonically decreasing curve. We compare aleatoric uncertainty from various layers, MC dropout, deep ensembles and the softmax entropy.

Our experiments show that aleatoric uncertainty from density estimates of deeper layers demonstrates similar calibration curves to established approaches. However, aleatoric uncertainty based on shallow layers can perform poorly (e.g. ResNet18 on SVHN). As discussed in section \ref{theory:pitfalls}, we argue that this results from difficulties in estimating the density of very high-dimensional hidden representations.

\subsubsection{Epistemic Uncertainty on \ac{ood} Detection}\label{experiments:ood_performance_epistemic}

\begin{table*}[t]
\footnotesize
\begin{center}
\setlength{\tabcolsep}{3pt}
\begin{tabular}{c l c c c c c c }
\toprule
 & OOD Data & L 1 & L 4 & L 7 & L 10 & Ensemble \\
 \midrule
 \multirow{6}{3mm}{\rotatebox{90}{\scriptsize Trained on MNIST}} & FashionMNIST & \textbf{0.975} & 0.922 & 0.855 & 0.811 & 0.896  \\
 & OMNIGLOT & 0.972 & 0.937 & 0.892 & 0.893 & \textbf{0.979}  \\
 & white noise & \textbf{1.000} & 0.972 & 0.903 & 0.841 & 0.785 \\
 & Rotated $90^{\circ}$ & \textbf{0.978} & 0.976 & 0.950 & 0.935 & 0.965 \\
 & HFlip & 0.902 & \textbf{0.907} & 0.883 & 0.864 & 0.905 \\
 & VFlip & \textbf{0.887} & 0.868 & 0.851 & 0.830 & 0.881  \\
 \midrule
 \multirow{6}{6mm}{\rotatebox{90}{\scriptsize \parbox{1.5cm}{Trained on\\FashionMNIST}}} & MNIST & 0.985 & \textbf{0.991} & 0.975 & 0.978 & 0.962 \\
 & OMNIGLOT & 0.971 & \textbf{0.987} & 0.960 & 0.967 & 0.960 \\
 & white noise & \textbf{1.000} & 0.985 & 0.971 & 0.930 & 0.840 \\
 & Rotated $90^{\circ}$ & \textbf{0.884} & 0.780 & 0.804 & 0.835 & 0.670 \\
 & HFlip & \textbf{0.719} & 0.696 & 0.701 & 0.693 & 0.657 \\
 & VFlip & 0.898 & 0.891 & 0.891 & \textbf{0.901} & 0.845 \\
 \bottomrule
\end{tabular}
\hfill
\begin{tabular}{c l c c c c c c }
\toprule
 & OOD Data & L 1 & L 9 & L 19 & L 34 & Ensemble \\
 \midrule
 \multirow{6}{3mm}{\rotatebox{90}{\scriptsize Trained on SVHN}} & CIFAR10 & \textbf{0.991} & 0.974 & 0.934 & 0.907 & 0.976 \\
 & STL10 & \textbf{0.999} & 0.991 & 0.951 & 0.912 & 0.982 \\
 & white noise & \textbf{1.000} & \textbf{1.000} & 0.986 & 0.903 & 0.992 \\
 & Rotated $90^{\circ}$ &0.615 & 0.646 & 0.689 & 0.918 & \textbf{0.957} \\
 & HFlip & \textbf{0.500} &\textbf{0.503} & \textbf{0.495} & \textbf{0.500} & \textbf{0.501} \\
 & VFlip & 0.506 & 0.520 & 0.551 & 0.708 & \textbf{0.736}\\
 \midrule
 \multirow{6}{3mm}{\rotatebox{90}{\scriptsize Trained on CIFAR10}} & SVHN & 0.042 & 0.029 & 0.091 & \textbf{0.736} & 0.723 \\
 & STL10 & 0.790 & \textbf{0.871} & 0.821 & 0.651 & 0.806 \\
 & white noise & \textbf{1.000} & \textbf{1.000} & \textbf{1.000} & 0.681 & 0.999 \\
 & Rotated $90^{\circ}$ & 0.553 & 0.517 & 0.543 & 0.757 & \textbf{0.824} \\
 & HFlip & \textbf{0.500} & \textbf{0.500} & \textbf{0.499} & \textbf{0.500} & \textbf{0.501} \\
 & VFlip & 0.519 & 0.513 & 0.537 & 0.714 & \textbf{0.789}\\
 \bottomrule
\end{tabular}
\vspace{-1mm}
\caption{\ac{ood} performance in terms of AUROC on \ac{ood} datasets when estimating epistemic uncertainty using the hidden activations. We evaluate the epistemic uncertainty computed from the distribution of the output of several affine layers. We evaluate layers (L) 1, 4, 7, 10 for a mulilayer perceptron trained on MNIST or FashionMNIST (\textbf{left}) and layers 1, 9, 19, 34 for a ResNet18 trained on SVHN or CIFAR10 (\textbf{right}). For comparison, we further report the performance of an ensemble of 10 identical networks. We observe that uncertainty estimates based on shallow layers demonstrate strong \ac{ood} performance.\vspace{-3mm}}\label{experiments:ood_performance}
\end{center}
\vspace{-3mm}
\end{table*}

This experiment evaluates the \ac{ood} detection performance of epistemic uncertainty. Table~\ref{experiments:ood_performance} reports average performance over 5 independent trainings, with standard deviations in the supplementary material. We measure the AUROC of the \ac{ood} detection using epistemic uncertainty of different layers and compare with deep ensembles.
When training on MNIST/FashionMNIST we evaluate on FashionMNIST/MNIST, OMNIGLOT \cite{lake2015human} and when training on CIFAR10/SVHN we evaluate on SVHN/CIFAR10, STL10 \cite{coates2011analysis}. Additionally we evaluate on test data perturbed with additive Gaussian noise, rotated by 90$^\circ$ and flipped horizontally/vertically. 

We generally observe a strong performance of epistemic uncertainty obtained from shallow layers which coincides with the results of section \ref{experiments:quality_of_uncertainty_depending_on_depth}, where we found that shallow layers yield more conservative estimates. However, we find that, when using a convolutional architecture (ResNet18), epistemic uncertainty obtained from shallow layers fails to detect \ac{ood} data generated by globally transforming the test data. We argue that this is expected behaviour since the hidden representations of shallow layers in convolutional networks denote low-level features that are likely to not represent such global transformations. Finally, we note the poor performance of shallow layers when training on CIFAR10 and evaluating on SVHN. These results suggest that recently found difficulties of explicit generative models on the task of \ac{ood} detection \cite{Nalisnick2018-gd} also apply to shallow layers. However, epistemic uncertainty from deeper layers does not suffer from this problem.

\subsection{Regression}\label{experiments:regression}

\begin{figure*}
    \centering
    \begin{subfigure}[t]{0.24\linewidth}
        \centering
        \vspace{0pt}
        \includegraphics[width=.65\textwidth]{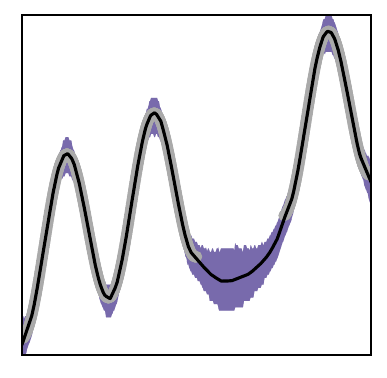}%
        \caption{Latent Density}
    \end{subfigure}
    \hfill
    \begin{subfigure}[t]{0.24\linewidth}
        \centering
        \vspace{0pt}
        \includegraphics[width=.65\textwidth]{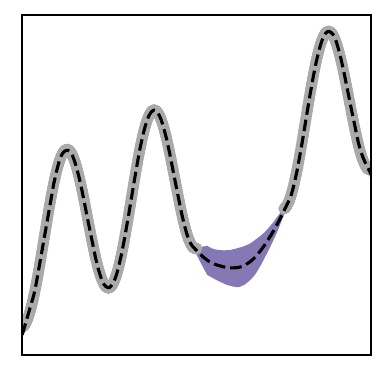}%
        \caption{Ensemble}
    \end{subfigure}
    \hfill
    \begin{subfigure}[t]{0.24\linewidth}
        \centering
        \vspace{0pt}
        \includegraphics[width=.65\textwidth]{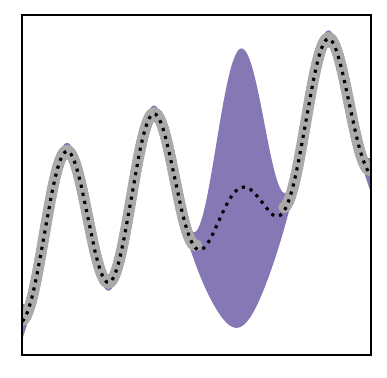}%
        \caption{Gaussian Process}
    \end{subfigure}
    \hfill
    \begin{subfigure}[t]{0.24\linewidth}
        \centering
        \vspace{0pt}
        \includegraphics[width=.7\textwidth]{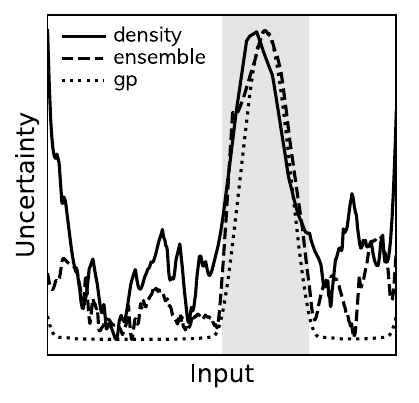}%
        \vspace{-2mm}%
        \caption{Epistemic Uncertainty}
        \label{fig:toy_regression_epistemic}
    \end{subfigure}%
    \vspace{-2mm}%
    \caption{Simple regression task. The models are trained on the data outlined in gray.
    \textbf{(a-c):} Model prediction is given in black and the high-confidence area in blue.
    In \textbf{(d)}, we compare the normalised estimated epistemic uncertainty of latent density (solid), deep ensemble (dashed) and a \ac{gp} (dotted). The gray background indicates the gap in the training data. Also for regression we find that the epistemic uncertainty estimated by density estimation matches established methods.
    \vspace{-5mm}}
    \label{fig:toy_regression}
\end{figure*}

As a simple regression experiment, we train regressors on data sampled from $f(x) = \frac{1}{2} \left( \sin (4 \pi x - \frac{\pi}{2}) + x \right)$ for $x \in \lbrack -1, 1\rbrack$\footnote{Designed to be naturally normalised to $[-1, 1] \rightarrow [-1, 1]$}. The training data and methods are illustrated in fig.~\ref{fig:toy_regression}. We train on the gray outlined data with a simple four-layer perceptron, extract embeddings at the penultimate layer (following our findings from section~\ref{experiments:image_classification}) and fit the density with a~\ac{cnf}. For inference, we perform numerical integration to evaluate the likelihood of the latent vectors. The same architecture is used for the ensemble. The \ac{gp} has a RBF kernel with parameters fit to the training data.

Fig.~\ref{fig:toy_regression} compares the latent density approach (a) to a Deep Ensemble (b) and a \ac{gp} (c). All methods show growing uncertainty with further distance to the training data, indicating that latent densities also contain information about epistemic uncertainty in regression networks. For a better evaluation, we compare $-\log p(\textbf{z}_i^\star)$ against other epistemic uncertainty methods in fig.~\ref{fig:toy_regression_epistemic}. We observe that all methods follow the expected behaviour of high uncertainty in the gray data gap and low uncertainty in the region of the training data, while the estimated epistemic uncertainty is slightly increasing towards the boundaries of the training distribution for the ensemble and the latent density.

\subsection{Effects of Regularisation}\label{experiments:reconstruction}
 To confirm that more informative hidden representations translate into better \ac{ood} detection, we apply an additional reconstruction loss (section \ref{theory:pitfalls}). We show this on FCNet trained on MNIST and evaluated FashionMNIST and on ResNet18 trained on SVHN and evaluated on STL10. We reconstruct the input from layer 10/19 (FCNet/ResNet18) \footnote{Unlike for FCNet we do not choose the last layer in tab. \ref{experiments:ood_performance} for ResNet18 since layer 34 has only 10 dimensions which we consider too small for reconstructing the input image.}. 

\begin{figure}
    \centering
    \vspace{0pt}
    \includegraphics[width=0.99\linewidth]{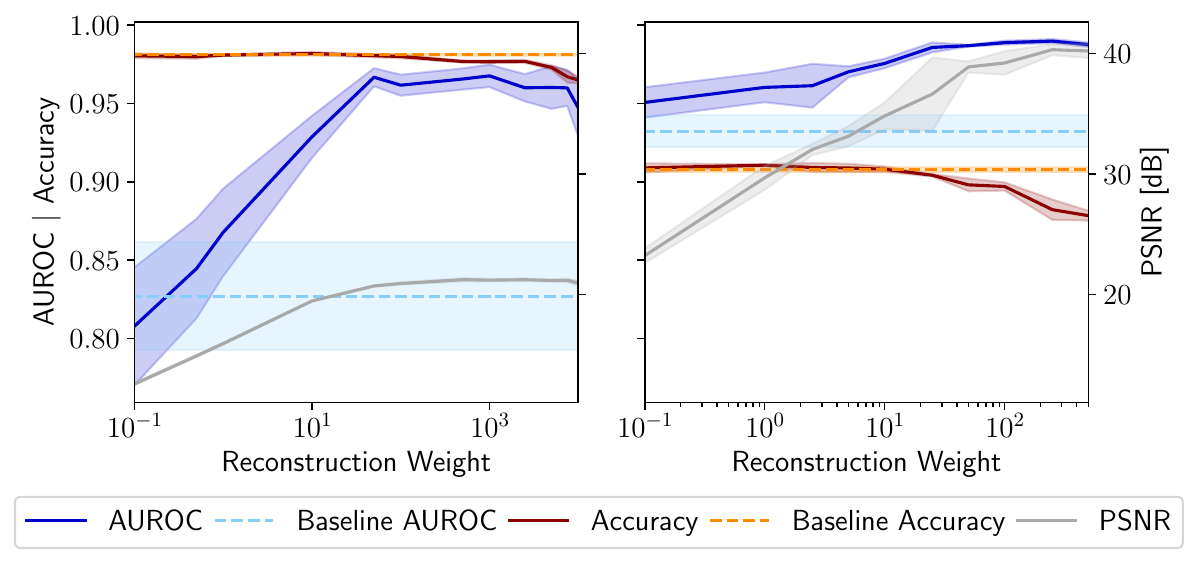}
    \label{fig:OODreconstruction}
    \vspace{-2mm}
    \caption{Detecting FashionMNIST/STL10 (left/right) when training on MNIST/SVHN with varying weight of the reconstruction loss. We use layer 10/19 for reconstruction and AUROC computation. Baseline Accuracy \& \ac{auroc} show the performance of a model trained without additional reconstruction loss. More informative representations boost \ac{ood} detection significantly.}
    \label{fig:OODmnist_reconstruction}
    \vspace{-3mm}
\end{figure}

Fig. ~\ref{fig:OODmnist_reconstruction} shows that in both cases the \ac{auroc} increases with the weight of the reconstruction loss and the reconstruction quality (PSNR). Most importantly, the improvement in \ac{auroc} is already achieved with low-weighted reconstruction losses that do not cause a noticeable reduction in accuracy.

\begin{figure}
    \begin{subfigure}[t]{0.99\linewidth}
        \centering
        \includegraphics[width=\textwidth]{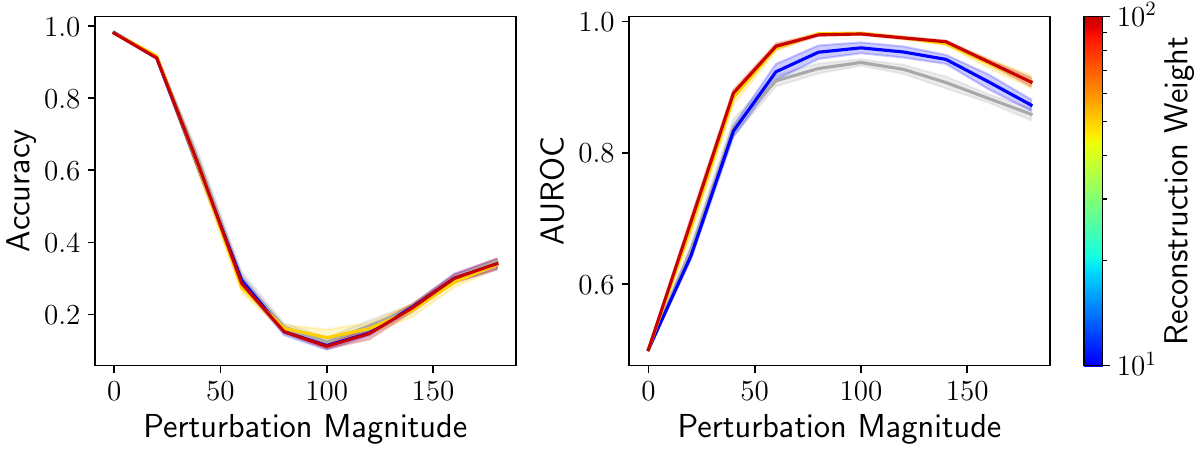}
    \end{subfigure}
    
    \begin{subfigure}[t]{0.99\linewidth}
        \centering
        \includegraphics[width=\textwidth]{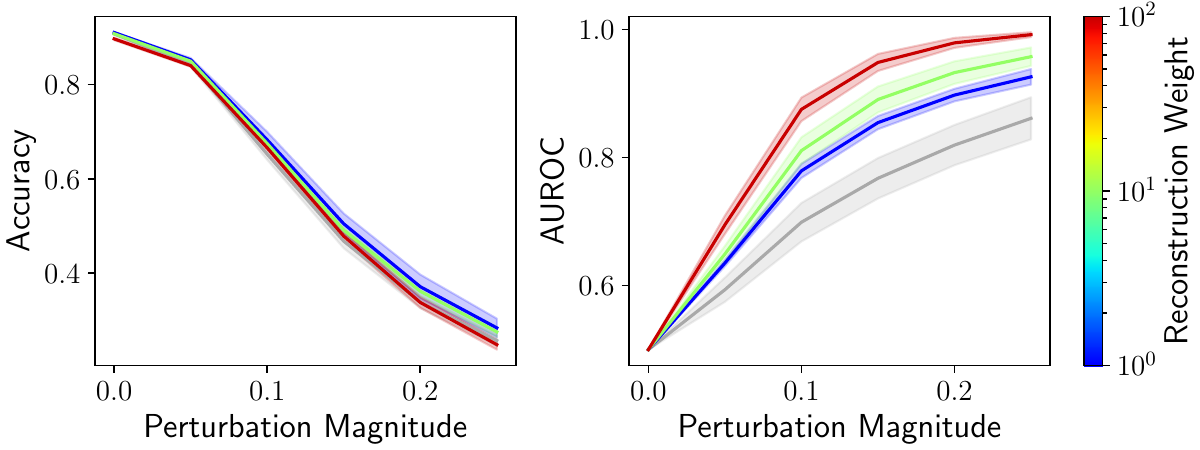}
    \end{subfigure}
    \vspace{-2mm}
    \caption{\ac{ood} detection performance on rotated MNIST (top) and SVHN with additive Gaussian noise (bottom) for various perturbation magnitudes and reconstruction weights. We reconstruct and estimate uncertainty from layers 10 (MNIST) and 19 (SVHN). The baseline without reconstruction loss is shown in gray. The AUROC increases with higher reconstruction losses while the Accuracy does not deviate much from the baseline.}
    \label{fig:OODreconstruction_perturbed}
    \vspace{-3mm}
\end{figure}

Fig.~\ref{fig:OODreconstruction_perturbed} shows how the accuracy and \ac{auroc} evolve when the original in-distribution dataset is perturbed with increasing magnitude\footnote{We choose the maximum reconstruction weight such it does not negatively impact best accuracy on the validation set.}. Following section \ref{experiments:image_classification} we apply rotations to MNIST and additive Gaussian noise to SVHN. We can see that the reconstruction weight has no impact on the shape of the accuracy curve, however we generally observe larger \acp{auroc} with increasing reconstruction weights. Note, that this ambiguity complicates the hyperparamter choice since it is difficult to determine which \ac{auroc} behavior close to the training data distribution is optimal. Training without the reconstruction loss tends induce strong correlation with the behavior MC dropout while higher weights tend to be close to deep ensembles (compare with fig. \ref{fig:qualitative_uncertainty}).

%
%
\section{Discussion \& Conclusion}\label{chapter:conslusion}%
This work introduced a holistic framework for uncertainty estimation based on the density of latent representations and verified its effectiveness empirically on the task of classification (section \ref{experiments:image_classification}) and regression (section \ref{experiments:regression}). In the absence of additional regularisation losses, we found that latent representations of shallow layers yield more conservative uncertainty estimates, while epistemic uncertainty obtained from deeper layers behaves similarly to MC Dropout and Deep Ensembles (section \ref{experiments:quality_of_uncertainty_depending_on_depth}). By applying perturbations of increasing magnitude to the test data (section \ref{experiments:quality_of_uncertainty_depending_on_depth}), we showed that latent densities denote an effective proxy for epistemic uncertainty and, thus, quantify a model's ability to generalise. This has promising practical implications since evaluating the latent density is faster than other state-of-the-art approaches and can be applied to any neural network post-training. Moreover, section \ref{experiments:quality_of_uncertainty_depending_on_depth} showed that the calibration of aleatoric uncertainty from deeper layers is similar to other well established methods.

Section \ref{experiments:ood_performance_epistemic} investigated the \ac{ood} performance of epistemic uncertainty and concludes that epistemic uncertainty from shallow layers has superior performance, which is in line with the results in section \ref{experiments:quality_of_uncertainty_depending_on_depth}. They even outperform Deep Ensembles for various \ac{ood} datasets, while adding negligible computational overhead and leaving the training procedure unchanged. This comes at the cost of worse uncertainty calibration close to the training data distribution (section \ref{experiments:image_classification}). We note that this result is achieved without applying an additional regularisation loss. However, we also find that certain global perturbations of the data distribution are better detected using the density of deeper layers. This suggests that the task of \ac{ood} detection can benefit from aggregating uncertainty information from several layers. We leave this exploration for future work. In fact, \cite{lee2018simple} utilizes density estimates from several layers. However, thus far they require logistic regression trained on \ac{ood} data for merging the results from several layers.

Moreover, section \ref{experiments:reconstruction} demonstrated that using a simple heuristic for increasing the information stored in the hidden representations further improves the \ac{ood} detection performance significantly, which is in line with prior work. However, our experiments in general showed that for uncertainty estimation \ac{ood} detection performance is not a sufficient metric. Different weights of the regularisation term lead to identical accuracy curves with very different \ac{auroc} curves close to the training data distribution. This leads to difficulties in choosing the optimal regularisation weight. Overall, if \ac{ood} detection is the primary objective of epistemic uncertainty additional regularisation can boost performance. However, note that a similar boost can also be achieved with density estimates of shallow layers while leaving the training procedure unchanged. When taking different aspects of epistemic uncertainty into account leaving out additional regularisation and using density estimates of deeper layers is a strong baseline given its simplicity and decent performance.

In conclusion, we find that the hidden activations of neural networks contain information about both aleatoric and epistemic uncertainty. Uncertainty proxies derived from density estimates of deep layers closely follow those of established - but computationally more expensive - methods. The finding that these uncertainties are available in already trained models enables a wide range of applications and rapid deployment.%
\bibliography{references}%
\bibliographystyle{icml2020}

\clearpage
\appendix
\section*{Appendix}\label{chapter:appendix}%

The supplementary material is structured in the following way: Section~\ref{app:background} provides additional explanations and proofs regarding the proposed method. Section~\ref{app:experiments} lists all experimental details and additional results to facilitate reproduction and reimplementaion of our experiments. Finally, section~\ref{app:realworld} shows two case-studies on deployment of latent activation based uncertainty estimation to real-world autonomous driving scenarios.

\section{Background}
\label{app:background}
\subsection{Process of Estimating the Output-Conditional Distribution of Hidden Representations}

Algorithm \ref{app:pseudocode} show schematically how to estimate the output-conditional density of hidden representations which is needed to quantify uncertainty with the proposed approach.

\begin{algorithm}[H]
\SetAlgoLined
 \KwIn{Dataset \{X, Y\}, Index l of layer used for uncertainty estimation, d maximum dimension of hidden representations.}
 \KwResult{Trained model $M$, output-conditional density model $M_{gen}$ of hidden representations at layer l, estimate of P($\hat{Y}$)} \BlankLine
 Train model $M$ on \{X, Y\} \BlankLine
 $Z_l$ $\leftarrow$ activations at layer l on \{X, Y\} \BlankLine
 $\hat{Y}$ $\leftarrow$ M(X) \BlankLine
 \If{$dim(Z_l) > d$}{
    $Z_l$ $\leftarrow$ reduce dimension of $Z_l$ using PCA} \BlankLine
    
 Initialize generative model $M_{gen}$ \BlankLine
 
 Train $M_{gen}$ on $Z_l$ given $\hat{Y}$ to estimate P($Z_l|\hat{Y}$) \BlankLine
 
 \eIf{classification}{
    Estimate marginal categorical distribution of network predictions P($\hat{Y}$) by counting frequencies \;
    }{
    Estimate marginal distribution of network predictions P($\hat{Y}$) with univariate parametric distribution (e.g. Gaussian, beta prime) \;}
 
 \caption{Estimating the output-conditional distribution of hidden representations.}\label{app:pseudocode}
\end{algorithm}

\subsection{Proof of Equation 3 and 4}
\label{app:proof}
Consider a deterministic neural network comprised of $L$ layers with $L-1$ latent representations $(\textbf{z}_0, .., \textbf{z}_{L-2})$ and some input $\textbf{x}$. Since the network is deterministic we know that $p(\textbf{z'}_i|\textbf{x}) = \delta(\textbf{z'}_i - f_i(\textbf{x}))$ and consequently: 

\begin{equation}
    \begin{aligned}
    p(\hat{\textbf{y}}|\textbf{x}) &=\int p(\hat{\textbf{y}}|\textbf{z'}_i)p(\textbf{z'}_i|\textbf{x}) d\textbf{z'}_i \\
    &=\int p(\hat{\textbf{y}}|\textbf{z'}_i) \delta(\textbf{z'}_i - f_i(\textbf{x}))d\textbf{z'}_i\\
    &= p(\hat{\textbf{y}}|\textbf{z}_i)
    \end{aligned}
\end{equation}
 where $\textbf{z}_i = f_i(\textbf{x})$ denotes the evaluation of the neural network until layer $i$. Eq. (3), $H(\hat{\textbf{y}}|\textbf{z}_i) = H(\hat{\textbf{y}}|\textbf{x})$ $\forall i\in [0, L-1]$, follows directly from that. Furthermore, according to the data processing inequality, any deterministic function $f$ applied to a random variable $x$ can only decrease its (differential) entropy: $H(x) \geq H(f(x))$. Considering that the $z_i$ are deterministic functions of the $z_{i-1}$ proves eq. (4): $H(\textbf{x}) \geq H(\textbf{z}_{0}) \geq ... \geq H(\textbf{z}_{L-2})$ $\forall i\in [0, L-1]$.

\subsection{Conditional Normalizing Flows}\label{app:cnf}
\ac{nfs}~\cite{dinh2014nice,dinh2016density} are a family of explicit generative models based on invertible neural networks. Assume we wish to model the distribution $p_{X}$ of some data $X$. Given some parametric distribution $p_{Z}$ and a NF $f_{\theta}$ comprised of weights $\theta$, likelihood evaluation of a data point $x \in X$ in \ac{nfs} is based on the change of variable formula
\begin{equation}
    p_{X}(x) = p_Z(f_{\theta}(x))\left| det\left( \frac{\partial f_{\theta}(x)}{\partial x} \right) \right|
\end{equation}
and made feasible by designing invertible layers that allow evaluating the determinant of their Jacobian efficiently. One such layer is the coupling layer introduced by~\cite{dinh2016density}. Each coupling layer splits its input activations into two sets $u_{1/2}^{in}$ and uses one set $u_1^{in}$ to compute scaling and translation of the other set $u_2^{in}$ using trainable, non-invertible functions $g_s$ and $g_t$, while it applies the identity function to $u_1^{in}$.

\begin{align*}
    u_1^{out} & = u_1^{in} \\
    u_2^{out} & = \left( u_2^{in} + g_t\left( u_1^{in} \right) \right) \odot g_s\left( u_1^{in} \right)
\end{align*}

This transformation renders the Jacobian a lower triangular matrix where the determinant is simply the product of the main diagonal. 

Several ways have been proposed to use \ac{nfs} for modeling the conditional distribution $p_{X|C}$ of $x \in X$ conditioned on some random variable $c \in C$. We adapt the conditioning scheme used in~\cite{Ardizzone2019-bv}, where the authors use a conditional version of the above coupling layer. Consequently, the coupling layers in the conditional normalizing flow $f_{\theta}(x|c)$ become:

\begin{align*}
    u_1^{out} & = u_1^{in} \\
    u_2^{out} & = \left( u_2^{in} + g_t\left( u_1^{in};c \right) \right) \odot g_s\left( u_1^{in};c \right)
\end{align*}

$g_s$ and $g_t$ are implemented as multi-layer perceptrons. To feed the conditional information into $g_s$ and $g_t$, we apply a separate multi-layer perceptron to $c$ and add the result to $g_t\left( u_1^{in}\right)$ and $g_s\left( u_1^{in}\right)$. 

\subsection{Estimating Aleatoric and Epistemic Uncertainty with Deep Ensembles and Monte-Carlo Dropout}\label{app:uncertainty_others}

In the case of MC dropout and deep ensembles, following \cite{gal2017deep} aleatoric uncertainty $u_{al}$ is computed as the average entropy of the predicted distribution (for several samples from the weight distribution): 

\begin{equation*}
    u_{al} = E_{w\sim p(w)} \left[ E_{y\sim p(\hat{y}|x, w)} \left[ \log(p(\hat{y}|x, w)) \right] \right]
\end{equation*}

Here, $p(\hat{y}|x, w)$ is the predicted distribution given an input $x$ and a set of weights $w$. $p(w)$ is the weight distribution (i.e. approximated posterior distribution). Here, we omit the conditioning of the weight distribution on the data. The epistemic uncertainty $u_{ep}$ is then estimated by approximating the mutual information between the weights $w$ and the predictions $\hat{y}$ conditioned on the input $x$:

\begin{align*}
    u_{ep} &= I\left( \hat{y}, w | x \right) \\ 
    &= H(\hat{y}| x) -  H(\hat{y}| w, x) \\
    &= E_{y\sim p(\hat{y}|x)} \left[ \log(p(\hat{y}|x)) \right] - u_{al}
\end{align*}

Here, $p(\hat{y}|x) = \int \mathrm{d}w p(w) p(\hat{y}|x, w)$ is evaluated using a finite set of samples/ensemble members.

\section{Experimental Details}
\label{app:experiments}
\subsection{Image Classification}

\subsubsection{Architectures}\label{app:image_architectures}

\begin{figure}
    \centering
    \begin{subfigure}[b]{0.8\linewidth}
        \centering
        \includegraphics[width=.9\linewidth]{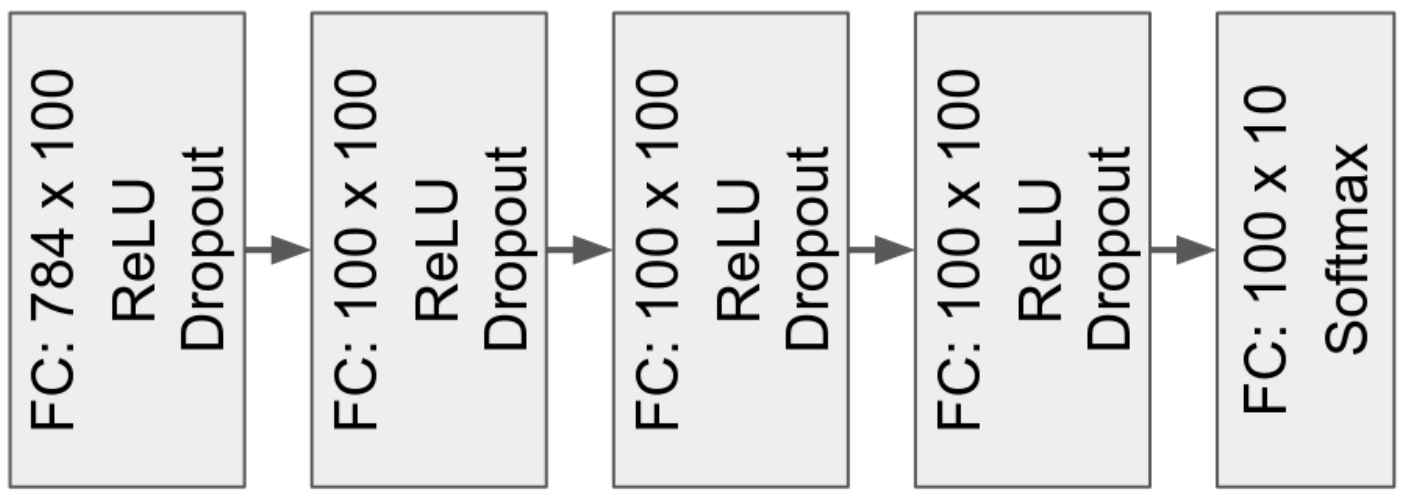}
        \caption{}
        \label{fig:regression_data}
    \end{subfigure}
    \caption{Architecture used in our image classification experiments. Multi-layer perceptron. FC denotes a fully-connected layer.}\label{fig:architectures_image_classification}
\end{figure}

The architectures of the neural networks used in our image classification experiments can be found in fig.~\ref{fig:architectures_image_classification}. We applied the multi-layer perceptron (fig.~\ref{fig:architectures_image_classification}) to MNIST and FashionMNIST and ResNet18 \cite{he2016resnet} to CIFAR10 and SVHN. Note that the dropout layer in FCNet only has $p>0$ when using MC dropout.

Furthermore, table \ref{app:idx2layer_fcnet} and \ref{app:idx2layer_resnet} show the correspondences between layer indices used in the main work and the layer type. Moreover, in the main work we primarily estimate the output-conditional distribution at the "Add" layers of ResNet18. This is done for convenience since   it allows estimating the density from a single layer output instead of taking to parallel branches of the computational graph into account.

\begin{table}
\begin{center}
\tiny
\begin{tabular}{ c c }
 Index & Layer \\ 
 0 & Flatten \\
 1 & Linear \\  
 3 & Dropout \\  
 4 & Linear \\  
 5 & ReLU \\  
 6 & Dropout \\  
 7 & Linear \\  
 8 & ReLU \\  
 9 & Dropout \\  
 10 & Linear \\  
 11 & ReLU \\  
 12 & Dropout \\  
 13 & Linear \\    
 14 & Softmax \\  
\end{tabular}
\end{center}
\caption{Layer index to layer type for FCNet.}
\label{app:idx2layer_fcnet}
\end{table}

\begin{center}
\begin{table}
\begin{center}
\tiny
\begin{tabular}{ c c }
 Index & Layer \\ 
 0 & Input \\
 1 & Conv2D \\
 2 & Batchnorm2D \\
 3 & ReLU \\
 4 & Conv2D \\
 5 & Batchnorm2D \\
 6 & ReLU \\
 7 & Conv2D \\
 8 & Batchnorm2D \\
 9 & Add \\
 10 & Dropout \\
 11 & ReLU \\
 12 & Conv2D \\
 13 & Batchnorm2D \\
 14 & ReLU \\
 15 & Conv2D \\
 16 & Conv2D \\
 17 & Batchnorm2D \\
 18 & Batchnorm2D \\
 19 & Add \\
 20 & Dropout \\
 21 & ReLU \\
 22 & Conv2D \\
 23 & Batchnorm2D \\
 24 & ReLU \\
 25 & Conv2D \\
 26 & Conv2D \\
 27 & Batchnorm2D \\
 28 & Batchnorm2D \\
 29 & Add \\
 30 & Dropout \\
 31 & ReLU \\
 32 & AveragePooling2D \\
 33 & Flatten \\
 34 & Linear \\
 35 & Softmax \\
\end{tabular}
\end{center}
\caption{Layer index to layer type for ResNet18.}
\label{app:idx2layer_resnet}
\end{table}
\end{center}

\subsubsection{Deep Ensemble and MC Dropout}

All deep ensembles in section \ref{experiments:quality_of_uncertainty_depending_on_depth} consist of ten neural networks of the same architecture as used by our approach. When training MC Dropout, we place an additional dropout layer after every dense layer (FCNet) or after every ResBlock (resnet).

\subsubsection{Training}

We describe the training of the neural networks on image classification and the output-conditional densities on the latent representations separately.

\textbf{Neural Network Training}

We train both, FCNet and ResNet18, using Adam \cite{kingma2014adam} ($\beta_1 = 0.9$, $\beta_2 = 0.999$) and a batch size of $32$. We use a learning rate of $0.001$ and weight decay of $0.0001$. We train the networks for $200$ epochs, while performing early stopping with a patience parameter of $20$. For all datasets, CIFAR10, SVHN, MNIST and FashionMNIST, we use fixed validation set containing $20\%$ of the training data. All reported results are on the test set of the corresponding dataset. 

\textbf{Output-conditional Density Training}

After training the neural networks, we extract the activations of the training data at layers of varying depth with the corresponding prediction of the neural network. Subsequently, we estimate the distribution of the latent representations of each class prediction with a separate \ac{gmm} of five components. We use the sklearn \ac{gmm} implementation\footnote{https://scikit-learn.org/stable/modules/generated/ sklearn.mixture.GaussianMixture.html}. Since the shallow layers of ResNet18 are high-dimensional ($\gg100$), density estimation - especially in the output-conditional case - becomes challenging. We counter this by reducing the dimensionality of the activations of each layer to 512 using a principal component analysis. Our multi-layer perceptron does not require dimensionality reduction, since each hidden layer has only 100 hidden units.

We also experimented with using \ac{nfs} for this task. However, we found that a \ac{gmm} is sufficient to fit the output-conditional distribution of latent representations throughout all layers.

\begin{table*}[t]
\footnotesize
\begin{center}
\begin{tabular}{c l c c c c c c }
\toprule
 & OOD Dataset & Layer 1 & Layer 9 & Layer 19 & Layer 34 & Ensemble \\
 \midrule
 \multirow{6}{3mm}{\rotatebox{90}{\scriptsize Trained on SVHN}} & CIFAR10 & $\textbf{0.991}\pm\textbf{0.001}$ & $0.974\pm0.004$ & $0.934\pm0.008$ & $0.907\pm0.003$ & $0.976\pm0.002$ \\
 & STL10 & $\textbf{0.999}\pm\textbf{0.000}$ & $0.991\pm0.002$ & $0.951\pm0.009$ & $0.912\pm0.003$ & $0.982\pm0.002$ \\
 & Gaussian noise & $\textbf{1.000}\pm\textbf{0.000}$ & $\textbf{1.000}\pm\textbf{0.000}$ & $0.986\pm0.011$ & $0.903\pm0.016$ & $0.992\pm0.002$ \\
 & Rotated ($90^{\circ}$) & $0.615\pm0.001$ & $0.646\pm0.010$ & $0.689\pm0.024$ & $0.918\pm0.002$ & $\textbf{0.957}\pm\textbf{0.002}$ \\
 & HFlip & $\textbf{0.500}\pm\textbf{0.000}$ & $\textbf{0.503}\pm\textbf{0.005}$ & $\textbf{0.495}\pm\textbf{0.010}$ & $\textbf{0.500}\pm\textbf{0.002}$ & $\textbf{0.501}\pm\textbf{0.002}$ \\
 & VFlip & $0.506\pm0.000$ & $0.520\pm0.004$ & $0.551\pm0.010$ & $0.708\pm0.006$ & $\textbf{0.736}\pm\textbf{0.001}$ \\
 \midrule
 \multirow{6}{3mm}{\rotatebox{90}{\scriptsize Trained on CIFAR10}} & SVHN & $0.042\pm0.001$ & $0.029\pm0.004$ & $0.091\pm0.016$ & $\textbf{0.736}\pm\textbf{0.055}$ & $0.723\pm0.048$ \\
 & STL10 & $0.790\pm0.010$ & $\textbf{0.871}\pm\textbf{0.019}$ & $0.821\pm0.038$ & $0.651\pm0.008$ & $0.806\pm0.004$ \\
 & Gaussian noise & $\textbf{1.000}\pm\textbf{0.000}$ & $\textbf{1.000}\pm\textbf{0.000}$ & $\textbf{1.000}\pm\textbf{0.000}$ & $0.681\pm0.057$ & $0.999\pm0.002$ \\
 & Rotated ($90^{\circ}$) & $0.553\pm0.003$ & $0.517\pm0.020$ & $0.543\pm0.005$ & $0.757\pm0.001$ & $\textbf{0.824}\pm\textbf{0.003}$ \\
 & HFlip & $\textbf{0.500}\pm\textbf{0.000}$ & $\textbf{0.500}\pm\textbf{0.000}$ & $\textbf{0.499}\pm\textbf{0.001}$ & $\textbf{0.500}\pm\textbf{0.002}$ & $\textbf{0.501}\pm\textbf{0.001}$ \\
 & VFlip & $0.519\pm0.003$ & $0.513\pm0.006$ & $0.537\pm0.006$ & $0.714\pm0.001$ & $\textbf{0.789}\pm\textbf{0.004}$ \\
 \bottomrule
\end{tabular}
\vspace{0.5pt}
\caption{\ac{ood} performance in terms of AUROC on various datasets when estimating epistemic uncertainty using the hidden activations of a resnet 18 on SVHN and CIFAR10. We evaluate AUROC using the epistemic uncertainty computed from the distribution of the output of several affine layers (1, 9, 19, 34). For comparison, we further report the performance of an ensemble containing 10 identical networks. We observe that uncertainty estimates based on shallow layers demonstrate strong \ac{ood} performance.\vspace{-3mm}}\label{experiments:ood_performance_resnet}
\end{center}
\vspace{-3mm}
\end{table*}

\begin{table*}[t]
\footnotesize
\begin{center}
\begin{tabular}{c l c c c c c c }
\toprule
 & OOD Dataset & Layer 1 & Layer 4 & Layer 7 & Layer 10 & Ensemble \\
 \midrule
 \multirow{6}{3mm}{\rotatebox{90}{\scriptsize Trained on MNIST}} & FashionMNIST & $\textbf{0.975}\pm\textbf{0.001}$ & $0.922\pm0.013$ & $0.855\pm0.032$ & $0.811\pm0.035$ & $0.896\pm0.018$ \\
 & OMNIGLOT & $0.972\pm0.002$ & $0.937\pm0.006$ & $0.892\pm0.006$ & $0.893\pm0.011$ & $\textbf{0.979}\pm\textbf{0.001}$ \\
 & Gaussian noise & $\textbf{1.000}\pm\textbf{0.000}$ & $0.972\pm0.005$ & $0.903\pm0.005$ & $0.841\pm0.010$ & $0.785\pm0.005$ \\
 & Rotated ($90^{\circ}$) & $\textbf{0.978}\pm\textbf{0.001}$ & $0.976\pm0.004$ & $0.950\pm0.005$ & $0.935\pm0.006$ & $0.965\pm0.002$ \\
 & HFlip & $0.902\pm0.002$ & $\textbf{0.907}\pm\textbf{0.005}$ & $0.883\pm0.007$ & $0.864\pm0.007$ & $0.905\pm0.001$ \\
 & VFlip & $\textbf{0.887}\pm\textbf{0.001}$ & $0.868\pm0.002$ & $0.851\pm0.004$ & $0.830\pm0.012$ & $0.881\pm0.002$ \\
 \midrule
 \multirow{6}{6mm}{\rotatebox{90}{\scriptsize \parbox{1.5cm}{Trained on\\FashionMNIST}}} & MNIST & $0.985\pm0.001$ & $\textbf{0.991}\pm\textbf{0.003}$ & $0.975\pm0.004$ & $0.978\pm0.005$ & $0.962\pm0.004$\\
 & OMNIGLOT & $0.971\pm0.001$ & $\textbf{0.987}\pm\textbf{0.002}$ & $0.960\pm0.012$ & $0.967\pm0.008$ & $0.960\pm0.003$\\
 & Gaussian noise & $\textbf{1.000}\pm\textbf{0.000}$ & $0.985\pm0.002$ & $0.971\pm0.006$ & $0.930\pm0.013$ & $0.840\pm0.004$\\
 & Rotated ($90^{\circ}$) & $\textbf{0.884}\pm\textbf{0.002}$ & $0.780\pm0.045$ & $0.804\pm0.030$ & $0.835\pm0.013$ & $0.670\pm0.013$\\
 & HFlip & $\textbf{0.719}\pm\textbf{0.003}$ & $0.696\pm0.007$ & $0.701\pm0.010$ & $0.693\pm0.008$ & $0.657\pm0.010$\\
 & VFlip & $0.898\pm0.006$ & $0.891\pm0.014$ & $0.891\pm0.006$ & $\textbf{0.901}\pm\textbf{0.013}$ & $0.845\pm0.014$\\
 \bottomrule
\end{tabular}
\vspace{0.5pt}
\caption{\ac{ood} performance in terms of AUROC on various datasets when estimating epistemic uncertainty using the hidden activations of a multilayer perceptron on MNIST and FashionMNIST. We evaluate AUROC using the epistemic uncertainty computed from the distribution of the output of several affine layers (1, 4, 7, 10). For comparison, we further report the performance of an ensemble of 10 neural networks of identical architecture. We observe that uncertainty estimates based on shallow layers demonstrate strong \ac{ood} performance. \vspace{-3mm}}\label{experiments:ood_performance_simple_fc}
\end{center}
\vspace{-3mm}
\end{table*}

\subsubsection{Error Margins}
We report the error margins over multiple runs on the \ac{ood} experiments in the tables~\ref{experiments:ood_performance_resnet} and \ref{experiments:ood_performance_simple_fc}.

\subsection{Regression}
\label{app:regression}
\subsubsection{Model Training}
For  the  regression prediction, we train a multilayer perceptron with 100 hidden units and 4 hidden layers until convergence. We then extract latent activations $z$ from the penultimate layer and train a \ac{cnf} conditioned on the predictions of the multilayer perceptron over the whole training data. The \ac{cnf} is based on GLOW~\cite{kingma2018glow} and consists of 4 layers with GLOW coupling blocks, each conditioned on the regression preduction, and random permutations. Since our training data is relatively small (750 samples), we do not use minibatches in both trainings.

For the ensemble, we take the same perceptron architecture and train 10 models with different random seeds. 

The GP has a simple RBF kernel and is fit (with  the sklearn implementation) to the same training data as the perceptron. The fit parameters are \texttt{0.601**2 * RBF(length\_scale=0.164)}.

\subsubsection{Evaluation}
To evaluate eq.~\ref{eq:epistemic}, we take a range of 1000 support points of $y \in [-10, 10]$ and estimate $p(\bm{z}|y)$ with the \ac{cnf}. We further assume a uniform prior $p(y)$ and numerically integrate over the support points to yield $\log p(\bm{z})$. 

For the visualised confidence regions, we start the integration at the prediction of the regression network and integrate separately in positive and negative direction until the integral reaches the set probability mass. In fig.~\ref{fig:toy_regression}, we visualize the integration boundaries to reach 20\% of the probability mass from the \ac{cnf}.

\subsection{Effects of Regularisation}

\subsubsection{Model Architecture}
\begin{figure}
    \centering
    \includegraphics[width=\linewidth]{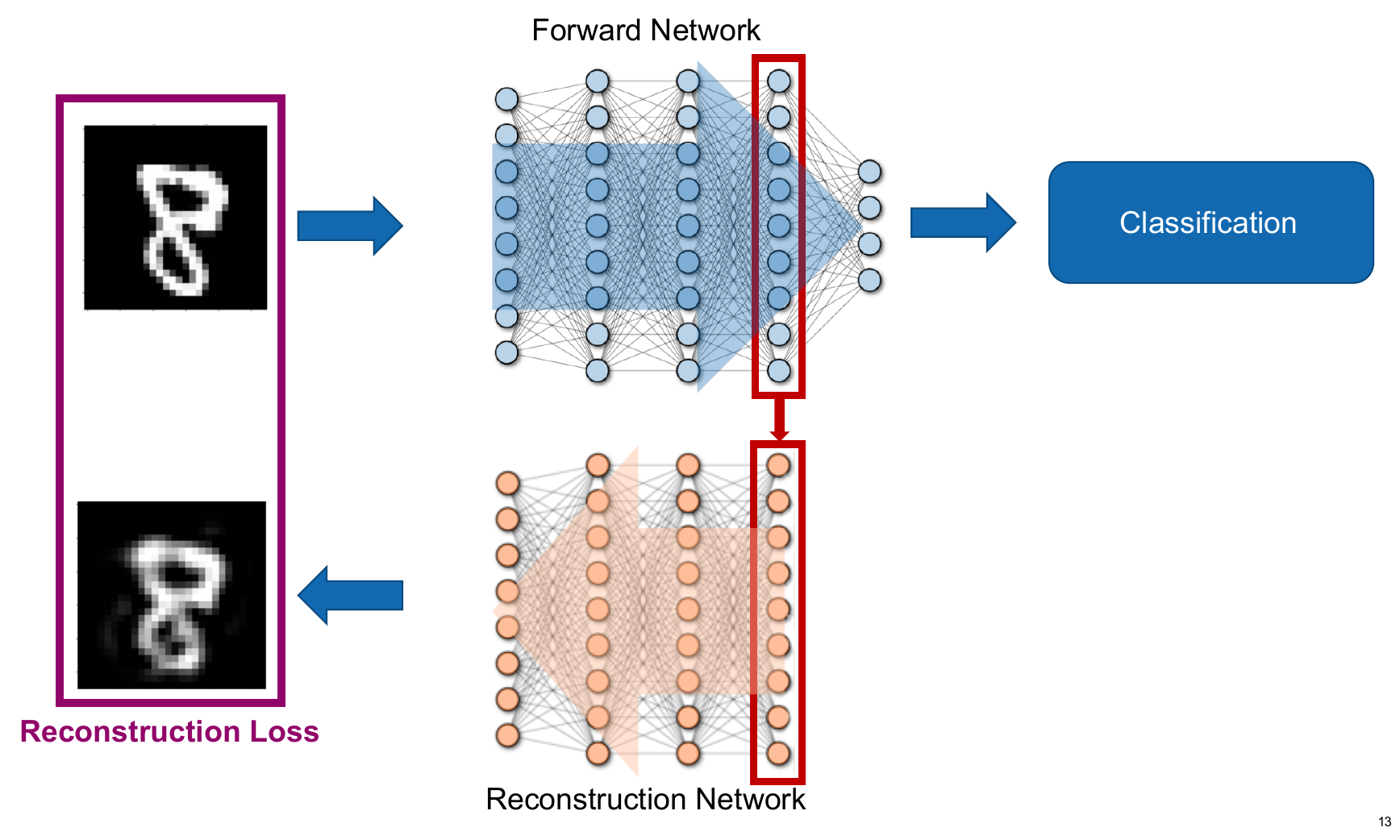}
    \caption{Overview of the model architecture with additional reconstruction network during training.}
    \label{fig:reconstruction_net}
\end{figure}

To explore the effects of regularization, we augment the model with a reconstruction network. The input to the reconstruction network is the latent vector extracted from the layer of our choice. The mapping from the input to the latent vector can be seen as the encoder. We choose the decoder for reconstruction to be symmetrical to the encoder such that their combination forms a symmetrical autoencoder. In the case of the fully connected network, we reconstruction the latent vector from layer $k$  using another  $k$ fully connected layers as shown in fig. \ref{fig:reconstruction_net}. The reconstruction network for the ResNet18 architecture follows a similar design. The difference is that the output of the convolutional layers has spatial dimensions, which we retain when feeding the latent tensor into the reconstruction network. The decoder is a ResNet18 model itself where the strided convolutions are replaced with a transpose convolution. We build the decoder with stacks of residual units where each stack upsamples the spatial resolution by a factor of two. The number of stacks can be directly inferred from the spatial resolution of the latent tensor.

\subsubsection{Model Training}
The combined model consisting of forward and reconstruction network are trained with the same parameters as described in section B.1.3. For fitting the output-conditional density after training, we only use the forward network. The reconstruction network is discarded after training as its sole purpose is regularization.
\section{Additional Experiments on Real-World Models}
\label{app:realworld}
In the following, we present two case studies on larger models that are trained to perform pixelwise predictions. Such models are often used in safety critical contexts like semantic segmentation for autonomous driving or segmentation of medical images. They also pose a specific implementation problem for density based models since the convolutional latent features cannot be accumulated accross the spatial dimension.

\subsection{Semantic Segmentation}
\label{experiments:semantic_segmentation}
As a large-scale classification experiment, we evaluate our method on semantic segmentation for urban driving. We use the state-of-the-art DeepLabv3+~\cite{chen2018deeplabv3} network trained on the Cityscapes urban driving dataset~\cite{Cordts2016Cityscapes} and estimate the feature density in the last layer before the logits.

\begin{table}[t]
    \centering
    \footnotesize
    \begin{tabular}{lr|rr}
\toprule
 & Misclass. & & OOD\\
 & Detection & \multicolumn{2}{r}{Detection}\\
 method & AUROC & AP & $\textrm{FPR}_{95}$\\
\midrule
softmax entropy & \textbf{94} & 2.9 & $45$\\
mc dropout & - & \textbf{9.8} & 38\\
\midrule
joint likelihood & 70 & 3.1 & \textbf{24}\\
conditional entropy & 92 & 0.8 & 100\\
\bottomrule
    \end{tabular}%
    \vspace{0.5cm}
  \caption{Comparison against softmax entropy and monte-carlo dropout on the two downstream tasks of misclassification and OOD detection in semantic segmentation. All values in [\%]. The aleatoric uncertainty estimated with the conditional entropy can detect misclassified pixel with high AUROC, matching the performance of the softmax entropy. For detecting anomalous objects, the epistemic uncertainty from mc dropout and the joint likelihood have very different performances in the two benchmark metrics. We conclude that aleatoric and epistemic uncertainty can be successfully extracted from latent activations and are separated as expected to indicate different uncertainty aspects.}
  \label{tab:segmentation}
\end{table}

\begin{figure}
    \centering
    \includegraphics[width=0.9\linewidth]{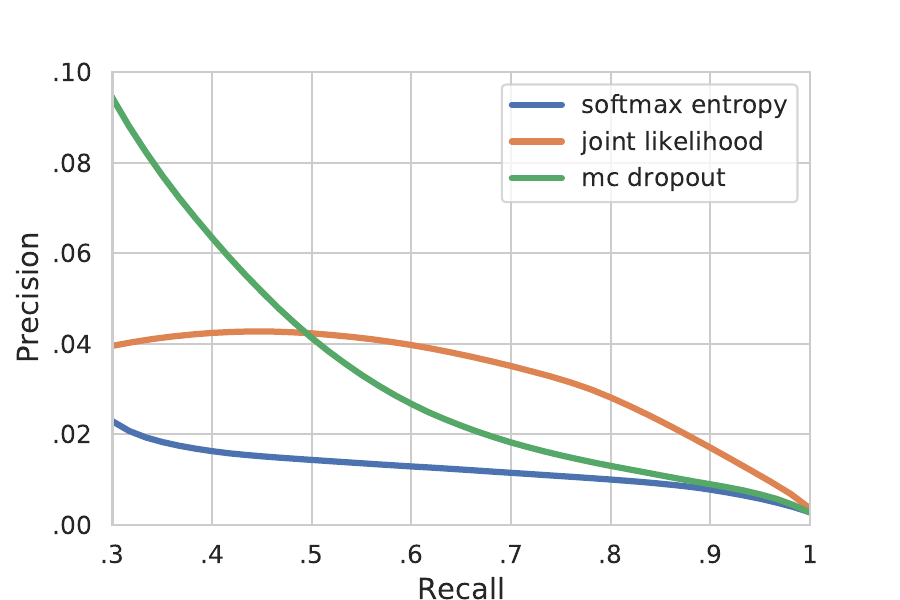}
  \caption{Precision-Recall curve for OOD detection in semantic segmentation on the Fishyscapes benchmark. The shown high-recall region is especially important for safety-critical applications. The curve illustrates the different characteristics that mc dropout and density based uncertainty exhibit on this task.}%
  \label{fig:segmentation}%

\end{figure}

\begin{figure*}
    \centering
    \def\iwidth{.25\linewidth}
\begin{tikzpicture}
\node[inner sep=0pt, label=\scriptsize{Input\strut}] at (0,0)
    {\includegraphics[width=\iwidth]{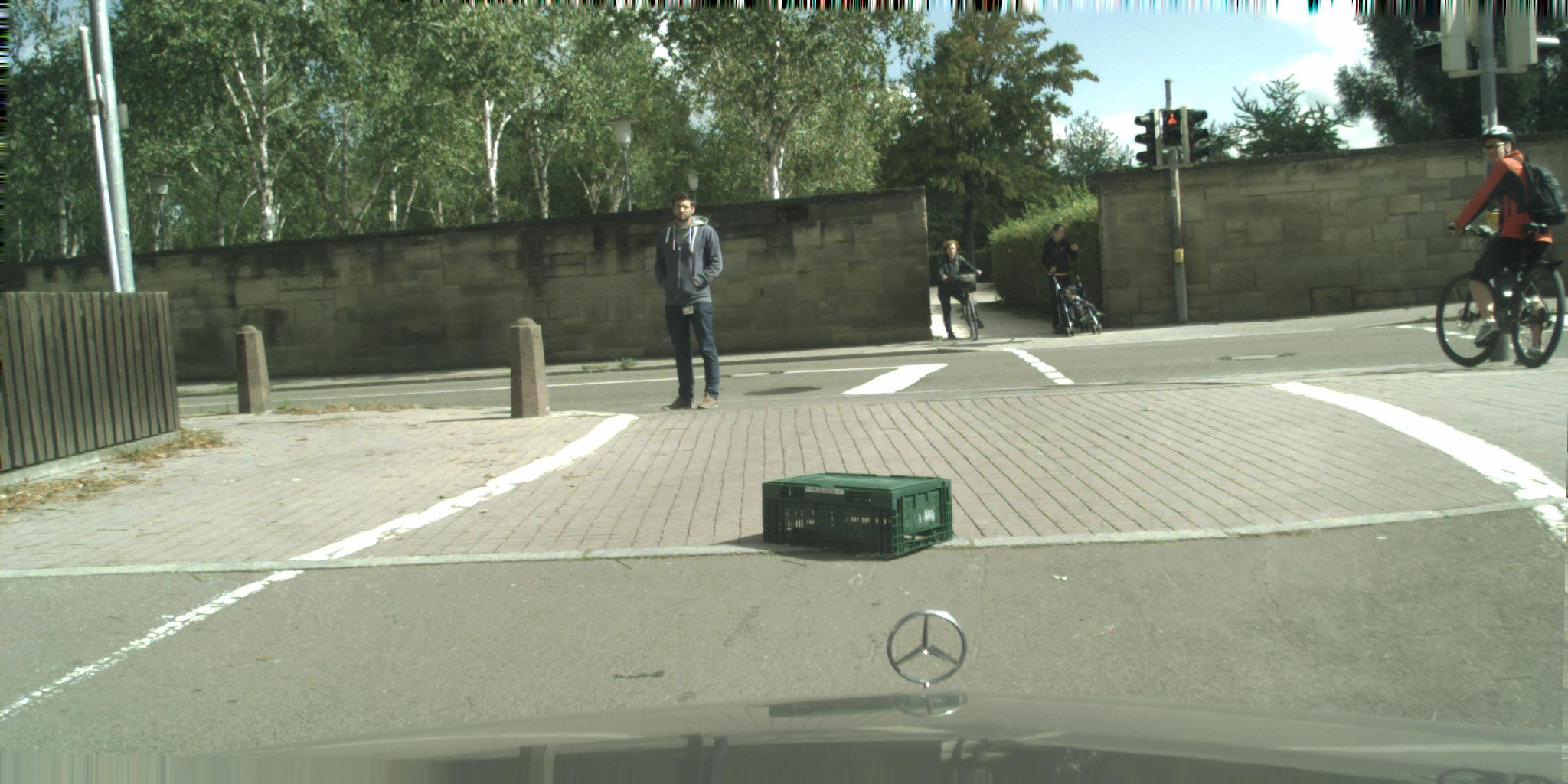}};
\node[inner sep=0pt, label=\scriptsize{Prediction\strut}] () at (\iwidth,0)
    {\includegraphics[width=\iwidth]{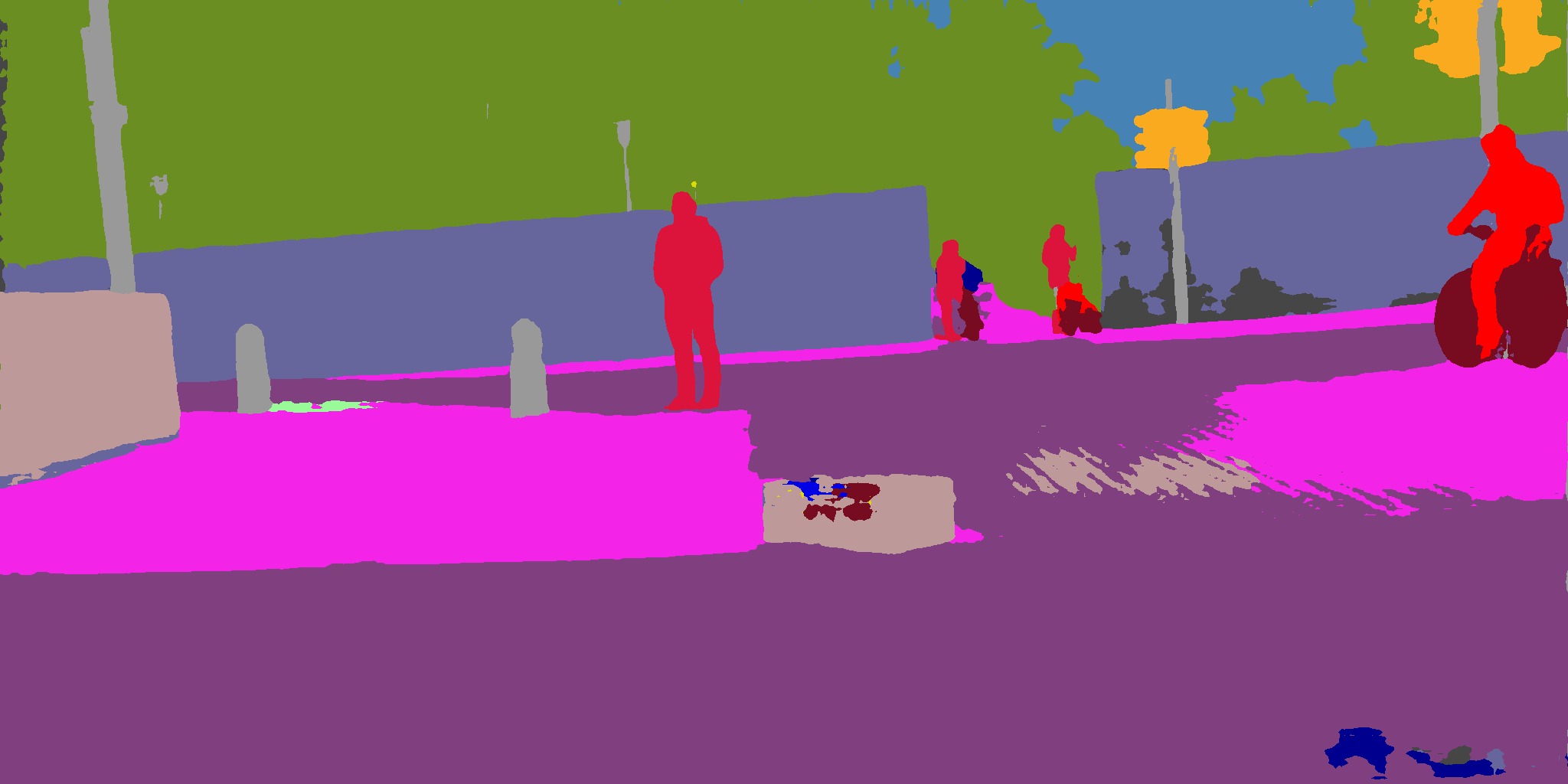}};
\node[inner sep=0pt, label=\scriptsize\strut Epistemic Uncertainty $\text{-}\log p(\textbf{z}^\star)$] () at (\iwidth*2,0)
    {\includegraphics[width=\iwidth]{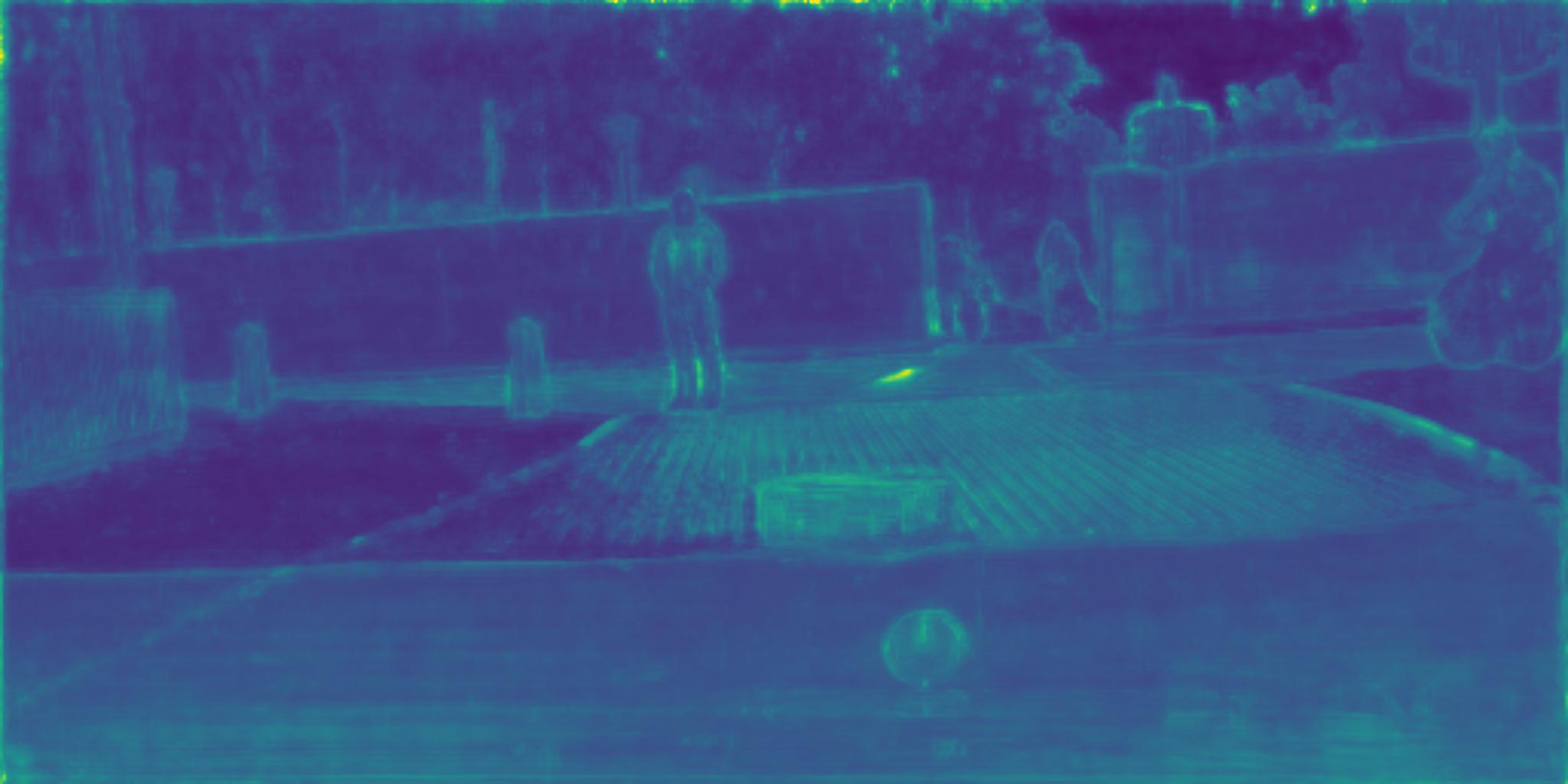}};
    
\node[inner sep=0pt] at (0,-2)
    {\includegraphics[width=\iwidth]{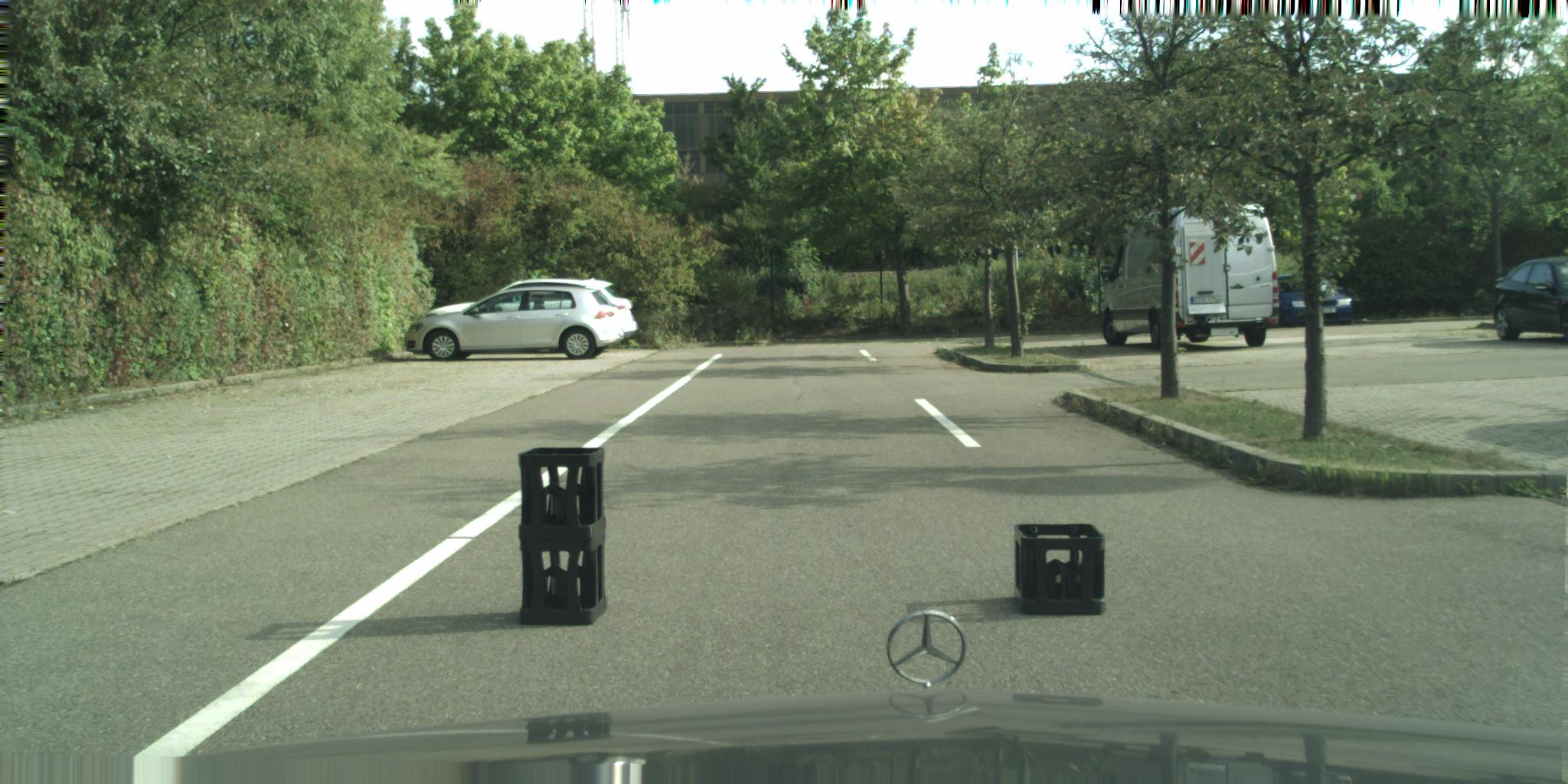}};
\node[inner sep=0pt] () at (\iwidth,-2)
    {\includegraphics[width=\iwidth]{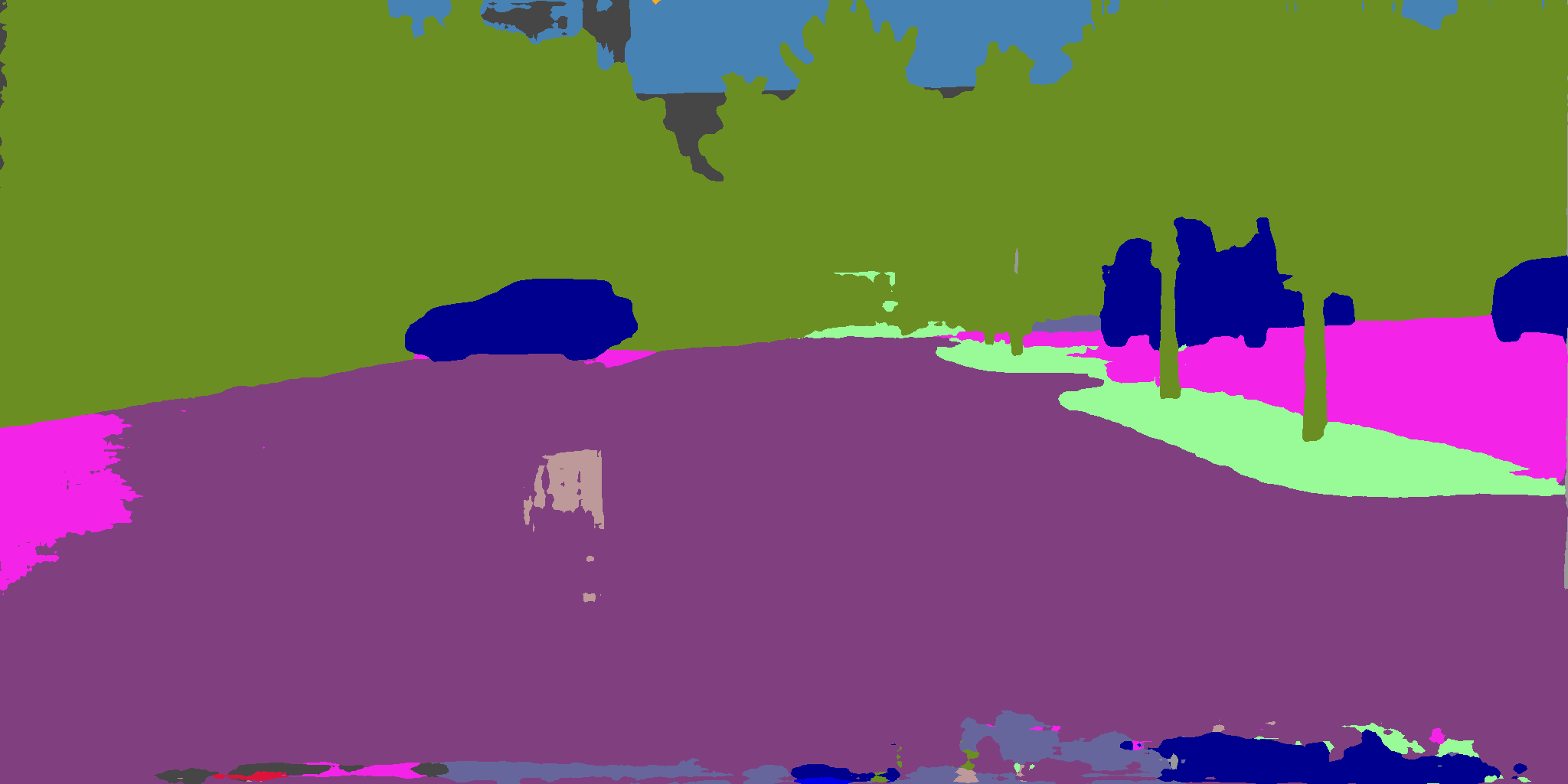}};
\node[inner sep=0pt] () at (\iwidth*2,-2)
    {\includegraphics[width=\iwidth]{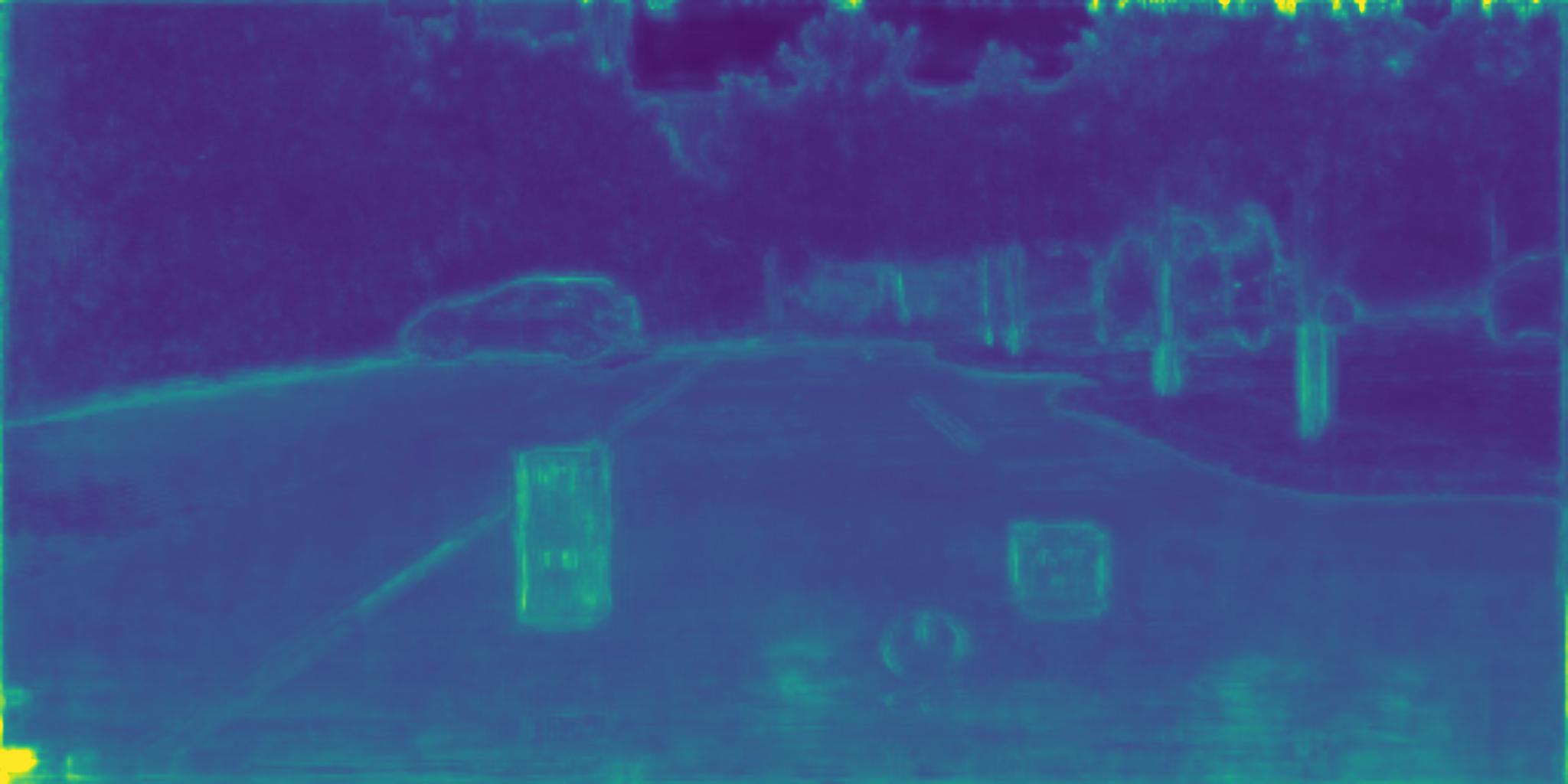}};
\end{tikzpicture}
    \caption{Qualitative examples of the estimated epistemic uncertainty in semantic segmentation. The examples are from the Lost \& Found dataset that features images similar to the training set, but with \ac{ood} objects~\cite{pinggera2016lost}. We observe that the estimated epistemic uncertainty is high for anomalous objects and the uncommon street texture in the first row and low where the model correctly generalised.}
    \label{fig:segmentation_examples}
\end{figure*}

\subsubsection{Experimental Setup}
We use the pre-trained DeepLabv3+~\cite{chen2018deeplabv3} provided by the authors\footnote{\url{https://github.com/tensorflow/models/blob/master/research/deeplab/g3doc/model_zoo.md\#\#deeplab-models-trained-on-cityscapes}} as the base network for the experiments with semantic segmentation. To train the output-conditional \acp{gmm}, we extract latent embeddings from the last layer before the logits (\texttt{decoder\_conv1\_0}) before the relu activation. We found the embedding dimensionality of 256 to be low enough to fit densities directly, without furhter dimensionality reduction. To associate latent vectors with predicted classes, we  interpolate them bilinearly to the output resolution of 1024x2048. To counter the correlations of latent vectors within the same image, we subsample a maximum of 500 vectors for every class and image. We then train the \acp{gmm} on a shuffled set of 100000 such vectors. We find suitable hyperparameters for the \acp{gmm} on the Fishyscapes Lost \& Found validation set and use the same hyperparameters for all classes. The best average precision on the validation set was achieved with 2 mixture components per class, tied covariance matrices, 1000 iterations and covariance regularisation factor $0.0001$.  We show qualitative examples in figure~\ref{fig:segmentation_examples}.

For estimating aleatoric uncertainty, we experienced numerical issues when evaluating $p(\hat{\textbf{y}}|\textbf{z}^\star)$. This is due to the normalization $\sum_y p(\hat{\textbf{y}}|\textbf{z}^\star) = 1$ when some classes are much more likely than others, i.e. when the entropy is low. We counteract these numerical issues by evaluating $h(\hat{\textbf{y}}|\textbf{z}^\star)$ on underfit \acp{gmm}\footnote{1 component, only 50 iterations, covariance regularisation with $0.1$} and on embedding vectors that were interpolated to output resolution (which was not found necessary for epistemic uncertainty and significantly increases memory usage).

\subsubsection{Results}
We test the estimation of aleatoric uncertainty by measuring the performance for misclassification detection on the Cityscapes validation data. Due to the nature of misclassifications, which are caused by the network itself, we can only directly compare against other methods that work on the exact same network. For \ac{ood} detection, we evaluate on the Fishyscapes anomaly detection benchmark~\cite{Blum2019-eh}. We report the results in table~\ref{tab:segmentation}.

We find that the aleatoric uncertainty estimated from output-conditional latent densities closely matches that of the softmax entropy. On the task of \ac{ood} detection the estimated epistemic uncertainty based on the log probability of the latent features does however not directly match the behaviour of the MC dropout. In particular, figure~\ref{fig:segmentation} shows a very bad performance in low-recall areas and a good performance for the high-recall area, contrary to MC dropout. The qualitative examples in figure~\ref{fig:segmentation_examples} show that outlier objects have a lower density, but there are also many other parts of the images with similar values. Further tests over different layers and architectures would be necessary to conclude whether this is an issue of e.g. feature collapse or poorly fitted densities.

Overall, we find that the holistic framework of aleatoric and epistemic uncertainty from latent densities scales with some adaptations to this complex task.

\subsection{Depth Regression}
\label{experiments:depth_regression}%

\begin{figure}
    \centering
    \includegraphics[width=\linewidth]{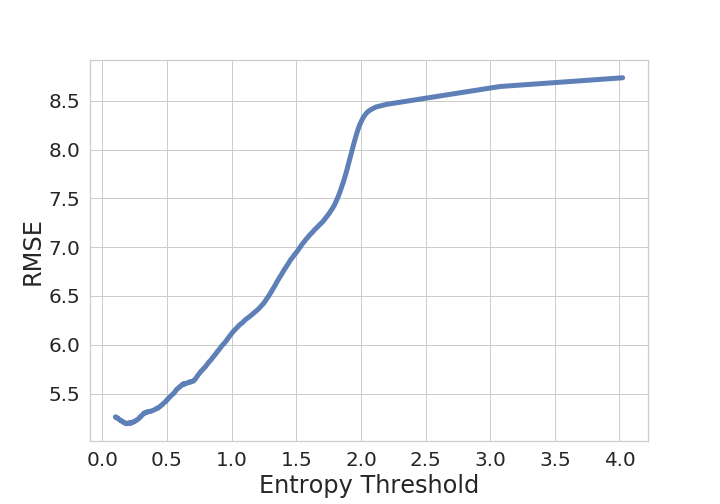}
    \caption{
    Average RMSE for a monocular depth prediction network \cite{godard2017unsupervised} trained on KITTI~\cite{Geiger2012CVPR}, accumulated below the given threshold of 
    aleatoric uncertainty $h(\hat{\textbf{y}}|\textbf{z}
   ^\star)$. As expected of aleatoric uncertainty, it increases monotonically for larger errors.\vspace{-5mm}}
    \label{fig:depth_regression_results}
\end{figure}

As a large scale application in regression, we apply our framework to monocular depth regression on KITTI \cite{Geiger2012CVPR}. 

\subsubsection{Experimental Setup}

We use a pretrained model from~\cite{godard2017unsupervised}\footnote{\url{https://github.com/mrharicot/monodepth}} trained on KITTI and extract the activations $z$ from the penultimate layer (output of the penultimate convolution) on the training data and the corresponding depth prediction. We need to estimate two densities: the output-conditional density $p(\textbf{z}^\star|\hat{\textbf{y}})$ and the density of the predictions $p(\hat{\textbf{y}})$. 

\textbf{Output-conditional Density of Latent Representations}

The conditional density $p(\textbf{z}^\star|\hat{\textbf{y}})$ is shared across all pixels. To assemble a dataset for training the density model on the training data, we extract the depth predictions pixel-wise  and all activations that are in the corresponding receptive field. This leads to a $144$ dimensional distribution conditioned on a scalar depth prediction. We randomly extract 12800 samples from the $256\times512$ dimensional depth maps and shuffle the resulting dataset to ensure that correlations between adjacent pixels have a minimal impact on the i.i.d. assumption when training the \ac{cnf}. Subsequently, we train a conditional version of the RealNVP~\cite{dinh2016density} on the resulting dataset, where we use a conditioning scheme similar to the one used by~\cite{Ardizzone2019-bv} (see section~\ref{app:cnf}). Our RealNVP consists of 3 coupling layers where scaling and translation are each computed using a separate multi-layer perceptron with 2 hidden layers of each 100 dimensions. We refer to the original work~\cite{dinh2016density} for a general description of the RealNVP. We train the CNF using Adam optimizer with learning rate $1e-6$ and weight decay $1e-5$. We further use a batch size of 128 and use early stopping with a patience parameter of $20$.

\textbf{Parametric Distribution of Depth Predictions}

\begin{figure}
    \centering
    \includegraphics[width=.5\linewidth]{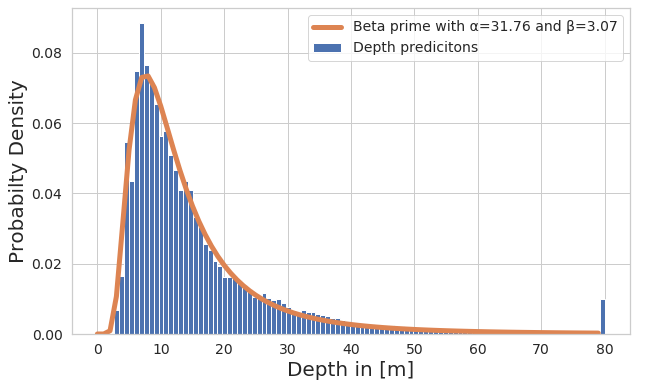}
    \caption{Fitting distribution of depth predictions with univariate beta prime distribution. We visualize normalized histogram of predicted depth values using a pretrained model from~\cite{godard2017unsupervised} and a beta prime distribution with parameters $\alpha =31.76$ and $\beta =3.07$. The small peak of the predicted depth values at 80m is an artifact of the code on which we base this experiment, since they clip depth predictions at 80m.}
    \label{fig:best_disitrbution_fit}
\end{figure}

We empirically found that the distribution of depth predictions on the training data is well represented by a univariate beta prime distribution with $\alpha =31.76$ and $\beta =3.07$. Fig.~\ref{fig:best_disitrbution_fit} shows the quality of this estimate.

\textbf{Evaluation of Aleatoric Uncertainty}

Evaluating  aleatoric uncertainty requires computing $h(\hat{\textbf{y}}|\textbf{z}^\star) = - \int d\hat{\textbf{y}} p(\hat{\textbf{y}}|\textbf{z}^\star) log \left( p(\hat{\textbf{y}}|\textbf{z}^\star) \right)$ where $p(\hat{\textbf{y}}|\textbf{z}^\star)=\frac{p(\textbf{z}^\star|\hat{\textbf{y}})p(\hat{\textbf{y}})}{p(\textbf{z}^\star)}$ and $p(\textbf{z}^\star)$ is obtained by marginalizing over $\hat{\textbf{y}}$. Since the depth predictions in our experiment are only one-dimensional, computing $h(\hat{\textbf{y}}|\textbf{z}^\star)$ is achieved in a straightforward manner using numerical integration. We discretize the interval of observed depth values $[3, 80]$ with 154 supporting points with equidistant spacing of $0.5$. Note that for higher-dimensional $\hat{\textbf{y}}$ above integral can be solved more efficiently using importance sampling.

\subsubsection{Results}

We evaluate the quality of the aleatoric uncertainty. We compute $h(\hat{\textbf{y}}|\textbf{z}^\star)$ according to eq.~\ref{eq:aleatoric} and evaluate its correlation with the RMSE in fig. \ref{fig:depth_regression_results}. As expected, larger errors have larger aleatoric uncertainty. We conclude that the aleatoric uncertainty estimate scales to large regression networks and allows extraction of aleatoric uncertainty post-training.

%

\end{document}